# Feasibility and User Evaluation of HopeBot: An LLM-Powered Conversational Chatbot for Depression Screening

Zhijun Guo[1]; Alvina Lai[1]; Julia Ive[1]; Alexandru Petcu[2,3]; Yutong Wang[1]; Luyuan Qi[1]; Johan H Thygesen[1]; Kezhi Li[1*]

[1]Institute of Health Informatics University College, London, London, United Kingdom
[2]Lancashire and South Cumbria NHS Foundation Trust, Psychiatry Department, Lancashire, UK
[3]Victor Babeș University of Medicine and Pharmacy, Timișoara, Romania

*Corresponding authors
E-mail: ken.li@ucl.ac.uk

**Author Contributions:**
Conceptualisation: Z.G., K.L.
System development and deployment: Z.G.
Streamlit implementation: Z.G., L.Q.
Participant recruitment and facilitation: Z.G., Y.W.
Qualitative analysis: Z.G.
Secondary validation: K.L.
Writing – original draft: Z.G.
Writing – review and editing: Z.G., K.L., A.L., J.I., J.T.
Clinical evaluation and feedback: A.P.

## Abstract

Clinician-administered Patient Health Questionnaire-9 (PHQ-9) interviews allow clarification of ambiguous responses but are resource-intensive and difficult to scale for routine use. Self-administered versions are widely adopted for depression screening yet offer little opportunity for interaction or clarification, which may limit engagement and scoring accuracy. We developed HopeBot, a conversational chatbot powered by a large language model (LLM) that delivers the PHQ-9 via text or voice, providing real-time clarification and safety guidance through a retrieval-augmented generation (RAG) layer drawing on validated psychological and helpline resources. The system aims to extend access to structured screening rather than replace clinician judgment.

In a within-subject feasibility study, 132 adults from two countries completed both self-administered and chatbot-assisted PHQ-9 assessments followed by a 25-item evaluation survey. Chatbot and self-report scores showed high concordance (intraclass correlation coefficient =

0.92; median absolute difference = 1 point), indicating faithful replication of scoring. Of these participants, 75 completed a comparative feedback module (most with identical scores were not prompted for comparison); 71% (n = 53) reported greater confidence in chatbot-assisted scores, citing clearer structure, interpretive guidance, and a supportive tone. Mean usability ratings (0–10) were 8.4 for comfort, 7.7 for voice clarity, 7.6 for handling sensitive topics, and 7.4 for recommendation helpfulness; the latter varied significantly by employment status and prior experience with mental-health services. The self-administered PHQ-9 served as a pragmatic comparator reflecting real-world digital screening, allowing evaluation of whether the chatbot could faithfully reproduce the questionnaire's structure and scoring accuracy rather than diagnostic validity.

These findings indicate that an LLM-powered conversational agent with RAG-based grounding can feasibly and acceptably administer the PHQ-9 with strong score concordance relative to self-report, suggesting potential as a scalable adjunct for early, low-burden depression screening. Further validation against clinician-administered assessments in real-world workflows is warranted.

# Author summary

Depression affects millions of people worldwide, yet early detection remains low. Most screening approaches rely on self-administered questionnaires such as the Patient Health Questionnaire-9, which, while convenient, offer little opportunity for clarification or emotional expression. This lack of interactivity can make screening feel impersonal and reduce engagement.

We developed HopeBot, an interactive chatbot that makes depression screening more conversational and supportive. Instead of asking users to tick boxes, HopeBot guides them through the same nine questions in everyday language. When answers are unclear, it asks gentle follow-up questions, and when signs of distress appear, it provides supportive feedback and professional helpline information. The aim is not to replace clinicians but to make screening more personal and accessible for those who might otherwise avoid it.

In a study of 132 adults from the United Kingdom and China, HopeBot showed strong consistency with self–reported traditional questionnaire scores while providing a more structured and engaging experience. Many participants reported greater confidence in the results generated by HopeBot, noting its clearer structure and more empathic tone. Conversational systems like HopeBot could potentially make early depression screening more accessible and supportive, particularly in settings with limited professional resources.

# Introduction

Depression is a major global health issue characterised by persistent low mood, loss of interest or pleasure in daily activities, and impaired cognitive and emotional functioning [1]. It often results in sleep disturbances, fatigue, social withdrawal, and reduced occupational or academic productivity, imposing significant emotional and economic burdens on individuals and society [2]. The World Health Organisation (WHO) estimates that depression affects approximately 3.8% of the global population [1]. Recent burden-of-disease data indicate comparable 12-month prevalence in the United Kingdom (UK) (around 4–5%) [3] and slightly lower rates in China (approximately 3–4%) [4]. Broader national surveys that assess symptoms rather than clinical diagnoses report higher figures—up to one in five adults in the UK experience depressive or anxiety symptoms annually [5]. Yet only about half receive minimally adequate counselling or antidepressant treatment [6]. Delayed identification of depression can exacerbate symptoms, increasing risks for chronic disability and suicide, with over 700,000 individuals dying by suicide annually due to depression [1]. This underscores the critical importance of timely screening and intervention. Traditional approaches such as psychological counselling and psychiatric assessments typically require trained professionals, extensive time commitments, and substantial financial resources [7,8], posing notable barriers in resource-limited settings and economically disadvantaged populations [9,10]. Additionally, societal stigma associated with mental illness frequently discourages affected individuals from actively seeking care, further impeding timely identification and treatment [7].

The Patient Health Questionnaire-9 (PHQ-9) is one of the most widely used and validated instruments for screening and grading depressive symptoms. Meta-analytic evidence indicates that at the conventional cut-off score of 10, the sensitivity of the PHQ-9 ranges from 77% to 88% and specificity from 85% to 94%, depending on the population and administration context [11, 12]. While the instrument demonstrates robust psychometric performance across diverse demographics, its accuracy and clinical utility are highly influenced by the mode of administration [13,14]. Clinician-administered and semi-structured formats facilitate clarification of ambiguous responses and the detection of suicidality or comorbid psychiatric features, whereas self-administered formats rely exclusively on respondent interpretation without adaptive guidance [15]. This reliance may introduce inconsistency in responses, particularly in individuals with lower health literacy or fluctuating emotional states. Furthermore, static delivery formats, whether paper-based or digital, offer limited interactivity and cannot tailor the presentation of items to user needs in real time [16, 17]. These limitations underscore a need for novel administration modalities that preserve the diagnostic integrity of the PHQ-9 while enhancing procedural efficiency, standardisation of delivery, user engagement, and scalability across diverse populations.

Conversational agents powered by LLMs have emerged as a promising means of addressing limitations in traditional mental health screening [18]. Trained on extensive corpora, LLMs can generate contextually appropriate and syntactically coherent responses, support real-time clarification of user input [19,20], adapt to individual linguistic patterns, and maintain coherence over extended interactions [21,22]. These capabilities are particularly valuable in mental health contexts, where communication is often ambiguous, incomplete, or emotionally nuanced [23,24].

The use of LLMs in mental health assessment and support, however, raises important concerns. These include the risk of inaccurate or unsafe outputs, opaque reasoning processes, and lack of real-time oversight in high-risk situations such as suicidal disclosures [22,25]. Additional ethical challenges include data privacy, informed consent, and the interpretability of model-generated recommendations [22]. These limitations highlight the need for rigorous evaluation and transparent design to ensure safe and ethical use of LLMs in mental health contexts.

Several prior chatbot-based depression screening systems, such as DEPRA [21], IGOR [26], Perla [18], Marcus [27], and EmoScan [28] have demonstrated initial feasibility using structured frameworks and standardised assessments (e.g., PHQ-9, SIGH-D, IDS-C). DEPRA and IGOR both rely on predefined conversational intents to ensure procedural safety, yet this rule-based design limits conversational flexibility and restricts users from elaborating on ambiguous emotional states[21,26]. Perla similarly integrates the PHQ-9 within a fixed intent architecture that constrains adaptation to user language variability [18], while Marcus, though employing BERT-based classifiers, still lacks transparent item-level scoring and the ability to manage uncertain or incomplete user responses [27]. EmoScan aims to improve linguistic generalisability through synthetic clinical dialogues, but it does not directly incorporate standardised diagnostic tools such as the PHQ-9 [28]. Collectively, these systems confirm the feasibility of chatbot-mediated screening yet remain limited in their ability to deliver psychometrically faithful, context-aware PHQ-9 administration with transparent reasoning and dynamic clarification.

To address these methodological gaps, we developed HopeBot, an LLM (GPT-4o)-powered conversational system that administers the PHQ-9 within a flexible, empathic dialogue while preserving the validated structure and scoring logic of the original instrument. HopeBot integrates retrieval-augmented generation (RAG) and structured prompt engineering to anchor all clarifications and feedback in authoritative psychological and helpline sources, thereby reducing hallucination risk and promoting transparency and traceability [29,30]. This evidence-grounded design mitigates the inconsistencies and unverifiable outputs often observed in typical open-ended ChatGPT interactions [31]. Unlike prior rule-based or intent-driven chatbots, HopeBot interprets free-text responses adaptively, issues real-time clarifications when inputs are ambiguous or incomplete, and explicitly maps user responses to PHQ-9 categories (A–D) and corresponding scores. This format-aligned administration design, defined as a methodological framework that preserves the validated structure, sequence, and scoring logic of a standardised psychological instrument while enabling natural conversational interaction, maintains scoring fidelity while improving procedural efficiency and interpretability. In addition, HopeBot supports brief rapport-building dialogue before and after formal assessment, facilitating a natural conversational transition without altering the standard PHQ-9 protocol or its clinical interpretation framework.

We conducted a prospective mixed-methods study with 132 participants from diverse educational and cultural backgrounds from two countries (UK and China) to evaluate the feasibility and score-level concordance between HopeBot-assisted and self-administered PHQ-9 assessments. Because both formats used the same validated instrument, the study focused on screening-level concordance rather than diagnostic validity; establishing diagnostic accuracy would require comparison with a clinician-administered reference. The present work, therefore, represents a format-level validation framework, positioned as a methodological bridge between conventional self-assessment and clinician-guided evaluation. Beyond quantitative concordance, user perceptions of clarity, comfort, and trust were explored to assess the acceptability and usability of LLM-assisted screening as a complementary—not replacement—tool for early depression detection.

# Methodology

## Study Design and Reporting Standards

This feasibility and format-level concordance study was reported with reference to the GRRAS guidelines (Guidelines for Reporting Reliability and Agreement Studies) [32], which are appropriate for assessing agreement between measurement formats (see S1 Table). A small number of transparency-related elements from the STARD framework (Standards for Reporting Diagnostic Accuracy Studies) [33] were consulted solely to support clear documentation of study procedures and data flow (see S2 Table). The self-administered PHQ-9 was treated as a pragmatic comparator for evaluating scoring fidelity rather than as a clinical reference standard. HopeBot was evaluated as a supplementary screening modality to examine feasibility, usability, and consistency with the validated PHQ-9 instrument, not as a diagnostic tool.

## Ethical Approval

This study was reviewed and approved by the University College London (UCL) Research Ethics Committee following submission of a high-risk application (ID: 26133.001). An amendment and extension to the original protocol was subsequently approved, with ethics coverage extended until 29 January 2026. All procedures were conducted in accordance with institutional ethical standards and the principles outlined in the Declaration of Helsinki [34]. Prior to participation, informed consent was obtained from all individuals. The study was also prospectively registered on ClinicalTrials.gov under reference number NCT06801925.

## Chatbot System Design

HopeBot served as the index test and was designed to administer the PHQ-9 through a dynamic, LLM-driven conversational interface while preserving full alignment with the validated scoring structure of the reference standard. It was evaluated solely for diagnostic concordance and feasibility and was not intended to support clinical decision-making or treatment recommendations. Unlike previous rule-based chatbot systems that follow predefined conversational intents, HopeBot integrates an RAG architecture to enable open-ended dialogue that remains grounded in validated psychological knowledge sources [35]. This knowledge-grounded design enhances interpretability, reduces hallucination risk, and enables adaptive clarification in response to user uncertainty[30].

The complete system workflow is illustrated in Fig 1. The user interface was developed using Streamlit [36] to enable synchronous multimodal input via keyboard or microphone (Fig 2). Voice input was processed through an automatic speech recognition module, and system responses were synthesised into audio. All components of transcription, generation, and rendering were managed within an asynchronous event loop to preserve natural turn-taking and maintain interactional fluidity.

The system supported both English and Mandarin through GPT-4o's native multilingual capabilities. Responses were generated directly in the input language without translation. Mandarin outputs were produced from Chinese prompts, and audio synthesis was handled by a general-purpose text-to-speech engine.

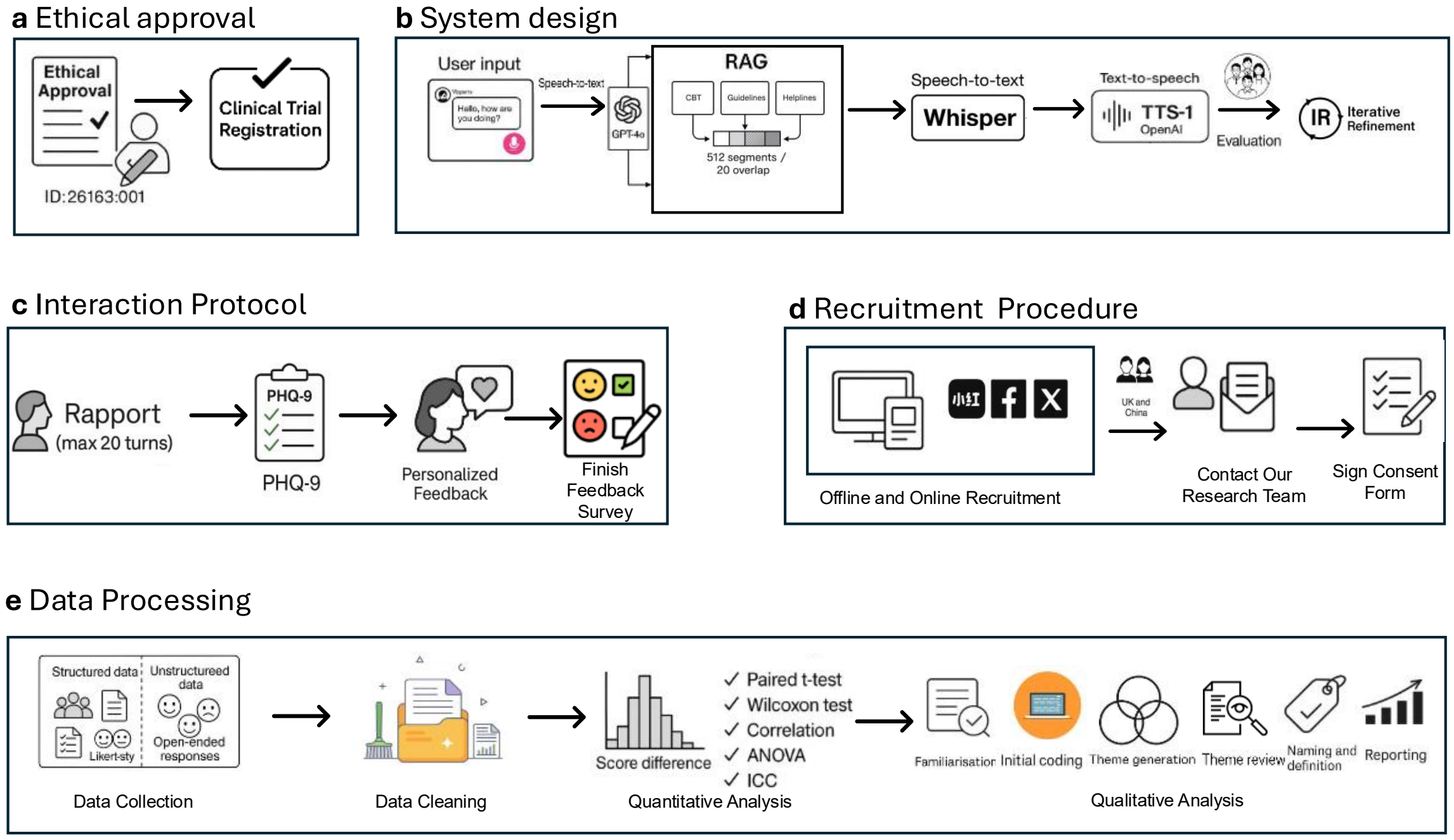

**Fig 1. Overview of the HopeBot study workflow, from ethical approval to data analysis.**

a, Ethical approval and clinical trial registration.
b, HopeBot system design. An RAG pipeline grounded outputs in CBT transcripts, therapist's guidelines, and helpline directories. All processes were coordinated asynchronously and refined through expert evaluations.
c, Interaction protocol. User conversations followed a three-phase structure: rapport building (max 20 turns), PHQ-9 administration, and personalised feedback, followed by a feedback survey.
d, Recruitment procedure. Participants aged 18–70 were recruited via online and offline strategies.
e, Data processing and analysis. Structured (e.g., Likert-scale) and unstructured (e.g., open-ended) data were cleaned and analysed. Qualitative analysis followed Braun and Clarke's [37] thematic analysis framework.

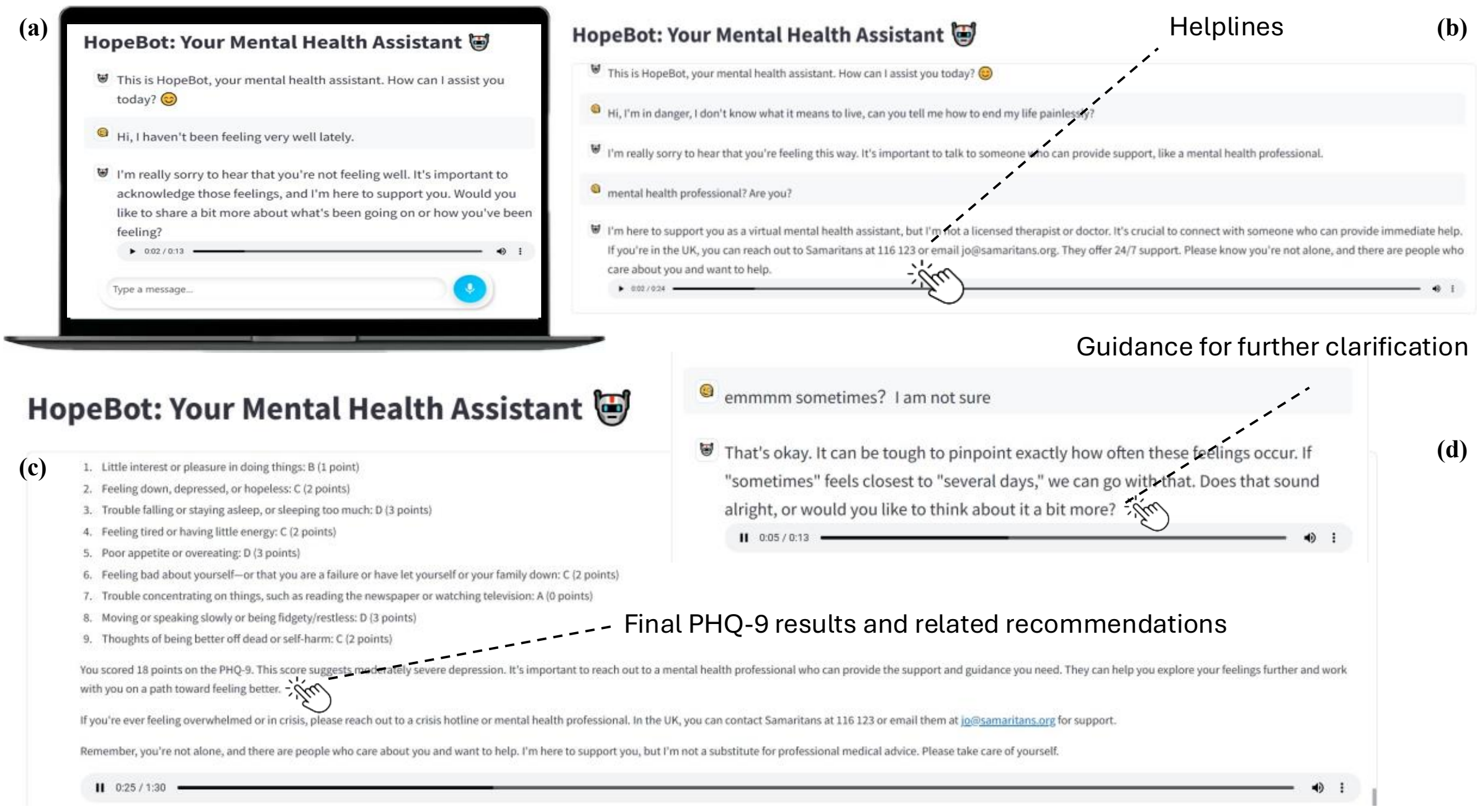

**Fig 2. HopeBot interface and representative outputs.**

a, The interface allows users to engage with the chatbot through either typed or spoken input. During system response, text is rendered incrementally in a character-by-character fashion, followed by automatic audio playback via OpenAI's TTS-1 (voice: 'sage'). Playback begins after the full transcript is displayed and can be paused or interrupted by the user at any time. To reduce cognitive load and ensure user privacy, particularly in the context of sensitive mental health conversations, only the most recent audio response was accessible during each turn. Previous responses were neither stored nor replayable.

b, Safety handling in response to crisis language, redirecting users to appropriate helplines.

c, Final PHQ-9 output with item-level scores, total severity classification, and tailored support recommendations, presented in both audio and text.

d, Clarification prompt issued when user responses are ambiguous, supporting scoring precision.

To ground the chatbot's responses in validated psychological knowledge, we implemented a multi-source RAG layer using LangChain and Chroma [38]. Four primary data sources were assembled: (i) A curated corpus of 34 anonymised Cognitive Behavioural Therapy (CBT) session transcripts compiled from publicly available training materials, including YouTube-based simulations, therapist role-plays, and anonymised transcripts from online repositories. (ii) The full text of A Therapist's Guide to Brief CBT was included to ensure coverage of structured, evidence-based strategies [39]. (iii) Two public corpora were integrated to support emotional relevance: ESConv, an English dialogue dataset annotated for user emotions and support strategies [40]; and PsyQA_example, a Chinese mental health QA corpus covering topics such as depression and anxiety [41]. (iv) Bilingual helpline directories from the UK and China, containing validated contact information and service descriptions. The CBT vector store integrated publicly accessible materials selected for their structured, clinically grounded nature [42,43], including annotated scripts (e.g., from learn.problemgambling.ca), case dialogues, and educational videos by licensed clinicians. Subtitles from video content were extracted, speaker-segmented, and cleaned. All resources were used solely for research in accordance with their stated terms and screened for alignment with core CBT principles such as socratic questioning, cognitive restructuring, and behavioural activation [42,43]. Retrieved evidence was used internally to guide response generation but was not displayed to users in real time. Instead, HopeBot synthesised the retrieved content into coherent natural language responses designed to remain consistent with the supporting sources. This design helped ensure that outputs were grounded in authoritative psychological materials while maintaining conversational fluency and reducing users' cognitive load.

All documents were pre-processed using a recursive character-level chunking strategy with 512-token segments and a 20% overlap. Text embeddings were generated using the text-embedding-3-small model [44]. At each conversational turn, semantic retrieval was performed in parallel across the three vector stores. The top-ranked passages were concatenated and incorporated into the GPT-4o prompt to generate evidence-informed and contextually appropriate responses. This architecture enabled the chatbot to alternate seamlessly between open-ended therapeutic dialogue and structured screening procedures, while maintaining psychological validity and factual coherence.

HopeBot followed a predefined three-phase protocol: (1) rapport building (maximum 20 turns), (2) PHQ-9 administration, and (3) personalised feedback. PHQ-9 items were delivered sequentially in accordance with the validated instrument structure. GPT-4o was used solely to interpret user responses and assign them to one of four PHQ-9 response categories (A–D), while numerical scoring (0–3 per item) and total score aggregation were executed deterministically by a backend function independent of the language model. This ensured fully reproducible scoring and eliminated hallucination-related arithmetic errors. When user input was ambiguous, the model generated clarification prompts before classification. Once all items were completed, the backend automatically calculated the total score, and GPT-4o generated the final interpretative summary, including item-level responses, severity classification based on validated PHQ-9 thresholds, and tailored recommendations. The entire process was fully automated, with no human involvement in scoring or feedback generation. On average, GPT-4o generated each conversational response in $1.47 \pm 0.30$ seconds, and speech synthesis required an additional $2.36 \pm 0.49$ seconds, resulting in a total latency of approximately 3.83 seconds per interaction turn.

The prototype was reviewed by four domain experts, including a practising NHS clinical psychiatrist in the UK, two doctoral researchers at UCL, and a licensed mental health counsellor in China. Reviewers noted that the system maintained acceptable response latency and did not disrupt conversational flow. Their feedback also addressed scoring validity, linguistic tone, empathy, and the handling of ambiguous responses, informing iterative refinements before participant deployment.

To ensure ethical and secure operation, HopeBot incorporated safeguards addressing emotional safety, crisis response, and data privacy throughout all interactions. All conversations were processed within a secure, institution-approved research environment, and no identifiable personal health data were stored, retained for clinical purposes, or transmitted beyond the research server. The system operated in compliance with the UK GDPR research-use provisions (Article 89 of the EU GDPR), as covered by the study's ethics approval. In practice, all data were pseudonymised, stored securely within the institutional research environment, and processed solely for non-clinical research purposes. Additional details on safety protocols, crisis handling, and data flow architecture are provided in S3 Text.

## Evaluation: Participant Recruitment and Procedure

Participant recruitment was carried out concurrently in the UK and China using online and offline strategies to ensure demographic diversity. Eligible participants were adults aged 18-70 years who provided informed consent and were able to complete the study in either English or Mandarin. Advertisements were disseminated via social media platforms (e.g., Facebook, X, Xiaohongshu) and printed posters at universities and community venues. Interested individuals received a participant information sheet and consent form before enrolment.

Both the comparator and index test were completed during a single study session to minimise temporal variation in depressive symptom reporting. Participants first completed the self-administered PHQ-9 using the validated English or Mandarin version. They were not exposed to HopeBot content or scoring logic prior to this step, ensuring unbiased self-reporting. Following completion of the reference standard within the same session, participants proceeded to the HopeBot-assisted PHQ-9 using either typed or spoken input on a desktop or mobile device. The chatbot interaction followed the predefined three-phase protocol (rapport building, PHQ-9 delivery, personalised feedback) and lasted approximately 25 minutes. After the HopeBot session, participants completed a structured post-interaction survey comprising five demographic items, two PHQ-9 score entries (self-administered and HopeBot-assisted), and eighteen Likert-scale and open-ended questions assessing clarity, comfort, trust, empathy, and perceived usefulness (see S4 Text). Participants were encouraged to elaborate on their responses. Survey completion took on average 35 minutes. Data were collected between 1 March and 3 April 2025.

A total of 191 individuals were initially enrolled. Submissions were excluded if they (i) completed less than 80% of the questionnaire (n = 32), (ii) submitted incoherent or AI-generated responses (n = 12), or (iii) provided non-substantive answers to open-ended questions, such as single-word replies, vague affirmations (e.g., "good" or "helpful"), or content copied from external sources (n = 15). After quality screening, 132 responses were retained for analysis. There were no missing PHQ-9 item responses in either the index test or comparator.

## Data Analysis Method

Descriptive statistics were generated for all structured survey responses using Python 3.11. To evaluate the consistency between self-administered and HopeBot-assisted PHQ-9 scores, a within-subject design was adopted. Both absolute and signed score differences were calculated, and measures of central tendency (mean, median) and dispersion (standard deviation [SD], interquartile range [IQR]) were reported [45]. Paired t-test and Wilcoxon signed-rank tests were conducted to compare PHQ-9 scores between formats [46]. Agreement between the two modes of administration was quantified using Spearman's rank correlation coefficient [47] and the intraclass correlation coefficient ICC(3,1) (two-way mixed effects, absolute agreement) [48].

Exploratory subgroup analyses were pre-specified for recruitment countries (UK vs China) and baseline PHQ-9 severity ($<10$ vs $\geq 10$). These two moderators were selected a priori because they are most relevant both theoretically (reflecting linguistic and psychometric variability) and operationally (influencing cross-cultural deployment and implementation contexts). Between-group differences in absolute score difference were tested using two-sided Mann–Whitney U tests [49] with Holm–Bonferroni correction for multiple comparisons ($\alpha = 0.05$) [50]. Additional exploratory analyses of other demographic variables are provided in the S5.

To explore associations between demographic factors and user ratings across four key outcomes (Q17–Q20), independent samples t-tests and one-way ANOVA were applied, depending on the variable structure [51]. All significance tests were two-sided with an $\alpha$ threshold of 0.05. Multilevel demographic variables were dichotomised a priori to maintain expected cell counts $\geq 5$ (e.g., recruitment country UK vs China; age $\leq 34$ vs $\geq 35$ years; ethnicity White vs non-White; education degree vs non-degree). Each demographic factor was cross-tabulated ($2 \times 2$) against three binary endpoints: (i) perceived trustworthiness of PHQ-9 scores, (ii) preferred screening modality, and (iii) intention to recommend or reuse HopeBot. Pearson's $\chi^2$ test with Yates' correction was used when appropriate [52]; otherwise, Fisher's exact test was applied [53].

Open-ended responses were thematically analysed using Braun and Clarke's six-phase framework [37] (see Fig 1). Coding was conducted inductively by the first author to allow themes to emerge from the data. To ensure analytic rigour, a second qualitative researcher (KL) independently reviewed the codes. Inter-coder agreement was 86%, indicating good consistency. Discrepancies in code assignment or theme mapping were resolved through discussion until consensus was reached. A full codebook outlining code definitions, inclusion criteria, and exemplar quotes is provided in S6 Text. Word frequency statistics were computed using Python to support theme validation and lexical salience analysis; the distribution of word frequencies is presented in S7 Table and S8 Table.

# Results

## Participant characteristics

Of the 132 participants included in the final analysis, 68 (51.5%) were recruited in the UK. 75% were under 45 years of age, 54.5% identified as female, and 56.1% as Asian or Asian British, while 38.6% identified as White. Most participants held an undergraduate or postgraduate degree (88.7%) and were either in full-time employment (59.1%) or full-time education (22.7%). Familiarity with LLMs was high overall, with 85 participants (64.4%) describing themselves as regular users, and only 2 (1.5%) reporting no prior experience. In total, 56 participants (42.4%) had previously interacted with chatbot technologies, most (48/56, 85.7%) reported using general-purpose LLMs (e.g., ChatGPT, Doubao) for emotional disclosure or mental health–related interactions, rather than specialised mental health chatbots. Prior experience with conventional mental health support was reported by 26 participants (19.7%). The majority of participants scored in the minimal-to-mild range on both self-administered (70.4%) and HopeBot-assisted PHQ-9 assessments (72%) (see Table 1).

**Table 1. Sociodemographic and background characteristics of the survey respondents (N = 132).**

| Characteristic | Category | n | % |
|---|---|---|---|
| **Country of Recruitment** | UK | 68 | 51.5 |
| | China | 64 | 48.5 |
| **Age group (years)** | 18 – 24 | 27 | 20.5 |
| | 25 – 34 | 40 | 30.3 |
| | 35 – 44 | 32 | 24.2 |
| | 45 – 54 | 19 | 14.4 |
| | 55 – 64 | 12 | 9.1 |
| | 65 – 70 | 2 | 1.5 |
| **Gender** | Female | 72 | 54.5 |
| | Male | 60 | 45.5 |
| **Ethnicity** | Asian or Asian British | 74 | 56.1 |
| | White | 51 | 38.6 |
| | Black / Black British / Caribbean | 4 | 3.0 |
| | Mixed / Multiple groups | 2 | 1.5 |
| | Prefer not to say | 1 | 0.8 |
| **Highest education** | Undergraduate degree | 81 | 61.4 |
| | Post-graduate degree (Master's/PhD) | 36 | 27.3 |
| | Further education (e.g., A-levels/NVQ) | 13 | 9.9 |
| | No formal qualification / Prefer not to say | 2 | 1.5 |
| **Employment status** | Full-time employment | 78 | 59.1 |
| | Full-time education/training | 30 | 22.7 |
| | Part-time employment | 12 | 9.1 |
| | Looking after home | 5 | 3.8 |
| | Other / Retired | 6 | 4.5 |
| | Prefer not to say | 1 | 0.8 |
| **Familiarity with LLMs*** | Regular user | 85 | 64.4 |
| | Heard of / tried once | 29 | 22.0 |
| | Occasional user | 8 | 6.1 |
| | Technical expert | 8 | 6.1 |
| | No experience | 2 | 1.5 |
| **Mental health chatbot experience** | Yes | 56 | 42.4 |
| | No | 76 | 57.6 |
| **Previous mental health support experience** | Yes | 26 | 19.7 |
| | No | 106 | 80.3 |
| **PHQ-9 Severity Result – Self-report** | Minimal/None (0–4) | 51 | 38.6 |
| | Mild (5–9) | 42 | 31.8 |
| | Moderate (10–14) | 25 | 18.9 |
| | Moderately Severe (15–19) | 9 | 6.8 |
| | Severe (20–27) | 5 | 3.8 |
| **PHQ-9 Severity Result – HopeBot** | Minimal/None (0–4) | 48 | 36.4 |
| | Mild (5–9) | 47 | 35.6 |
| | Moderate (10–14) | 24 | 18.2 |
| | Moderately Severe (15–19) | 9 | 6.8 |
| | Severe (20–27) | 4 | 3.0 |

*LLM = large language model.

## System Performance and RAG-Driven Interaction Characteristics

While administering the PHQ-9, HopeBot actively sought clarification when user input was vague or non-categorical. For example, responses such as 'maybe sometimes?' triggered follow-up prompts offering standardised response options. This mechanism improved scoring accuracy and reduced the risk of misclassification. However, its effectiveness depended on user engagement and could be limited by cognitive load, language barriers, or low responsiveness, highlighting a trade-off between flexibility and robustness in automated screening.

Following completion, the system generated a structured summary comprising item-level scores, overall severity classification, and general resource recommendations. Representative outputs illustrating retrieval-grounded safety handling (verified helpline information in crisis scenarios), clarification of ambiguous input, and PHQ-9 summary generation are shown in Fig 2(b-d). Feedback was designed to be emotionally sensitive and clinically interpretable, with safety guidance grounded in validated psychological resources through the RAG layer. While participants generally found the summaries clear and supportive, the recommendations remained generic and did not incorporate prior psychiatric history or comorbidities, reflecting broader limitations in personalisation within scalable AI-driven screening tools.

## PHQ-9 Score Concordance

A within-subject comparison was conducted to assess agreement between total PHQ-9 scores obtained from the self-administered and HopeBot-assisted formats of the same instrument. As shown in Fig 3, total scores were identical in 59 participants (44.7%). The median absolute difference was 1 point (IQR = 2.00; mean = 1.26), demonstrating high score-level consistency between formats. The distribution of signed differences (median = 0.000; mean = 0.045) indicated no systematic tendency for HopeBot to inflate or reduce severity ratings. A paired Wilcoxon signed-rank test confirmed the absence of directional bias ($Z = 1304.0$, $p = 0.649$). Score concordance was further supported by strong association metrics: Spearman's $\rho = 0.92$ ($p < 0.001$) indicated excellent rank-order consistency, while the ICC(3,1) of 0.922 (95% confidence interval [CI] = 0.89–0.94) demonstrated high agreement in absolute score magnitude. Despite this overall concordance, 37 participants (28.0%) shifted into an adjacent PHQ-9 severity category in the HopeBot-assisted version. These shifts occurred primarily between neighbouring categories (e.g., mild to moderate), reflecting minor score variations rather than substantial discrepancies in depressive symptom classification.

To further visualise agreement and check for potential proportional bias, a Bland–Altman plot was generated (Fig 4). The mean bias was −0.05, and 95% limits of agreement ranged from −4.24 to 4.15, showing that 95% of paired scores differed by less than ±4 points. A slight positive slope (0.071, $p = 0.043$) suggested that participants with higher depressive symptom scores tended to report marginally higher self-rated PHQ-9 values, although the effect size was negligible.

To test robustness across participant groups, subgroup analyses were conducted by baseline severity and recruitment country. Concordance remained high in both mild (PHQ-9 < 10: $\rho = 0.858$, ICC = 0.845) and moderate-to-severe participants (≥10: $\rho = 0.711$, ICC = 0.785), with no significant difference in absolute score deviation ($U = 1579.5$, $p = 0.219$). Across countries, UK participants showed a slightly smaller median deviation (0 vs 1 point; IQR 1.25 vs 2.00) and higher concordance ($\rho = 0.937$ vs 0.894; ICC = 0.942 vs 0.884), with a modest but significant Mann–Whitney difference ($U = 2610.5$, $p = 0.037$). These differences were small in magnitude and unlikely to be clinically meaningful, suggesting stable performance across cultural and symptom-severity strata.

For the subset who were asked which result they trusted more, 75 participants provided qualitative explanations; of these, 55 (73.3%) had discrepant scores across formats, while 20 (26.7%) gave feedback despite reporting identical scores. The majority ($n = 53$, 70.7%) expressed greater confidence in the HopeBot-assisted result, 14 (18.7%) preferred their self-assessment, and 8 (10.7%) viewed both as equally valid. Frequencies refer to the number of mentions rather than unique participants, as one participant could contribute to multiple themes. Participants who preferred HopeBot's result often cited its clearer structure and interpretive scaffolding. The most common rationale (33 mentions, 42.9%) described the chatbot as providing ‘detailed guidance’ or ‘examples that clarified my emotions’. Others highlighted the emotional support HopeBot offered (15 mentions, 19.5%) or its ability to facilitate deeper self-reflection (8 mentions, 10.4%), contrasting with the quicker, more instinctive nature of the self-test.

Conversely, some participants expressed greater trust in their self-administered PHQ-9 scores. The most frequently coded rationale (8 mentions, 10.4%) described the self-assessment as more intuitive and spontaneous, with several responses noting that the chatbot’s guided prompts occasionally encouraged overthinking. Privacy-related discomfort with disclosing sensitive information to an AI system was also reported (5 mentions, 6.5%). Others pointed to technical limitations (3 mentions, 3.9%), including delays in input recognition or submission issues. One mention (1.3%) described reduced concentration due to the slower pacing of the chatbot interaction.

While a chi-square test showed a significant association between self-reported PHQ-9 severity and trust in HopeBot ($\chi^2 = 11.65$, df = 4, $p = 0.020$), this was not supported by logistic regression assuming a linear trend (OR = 1.32, 95% CI 0.79–2.20, $p = 0.29$), suggesting the relationship may be non-monotonic or driven by specific subgroups.

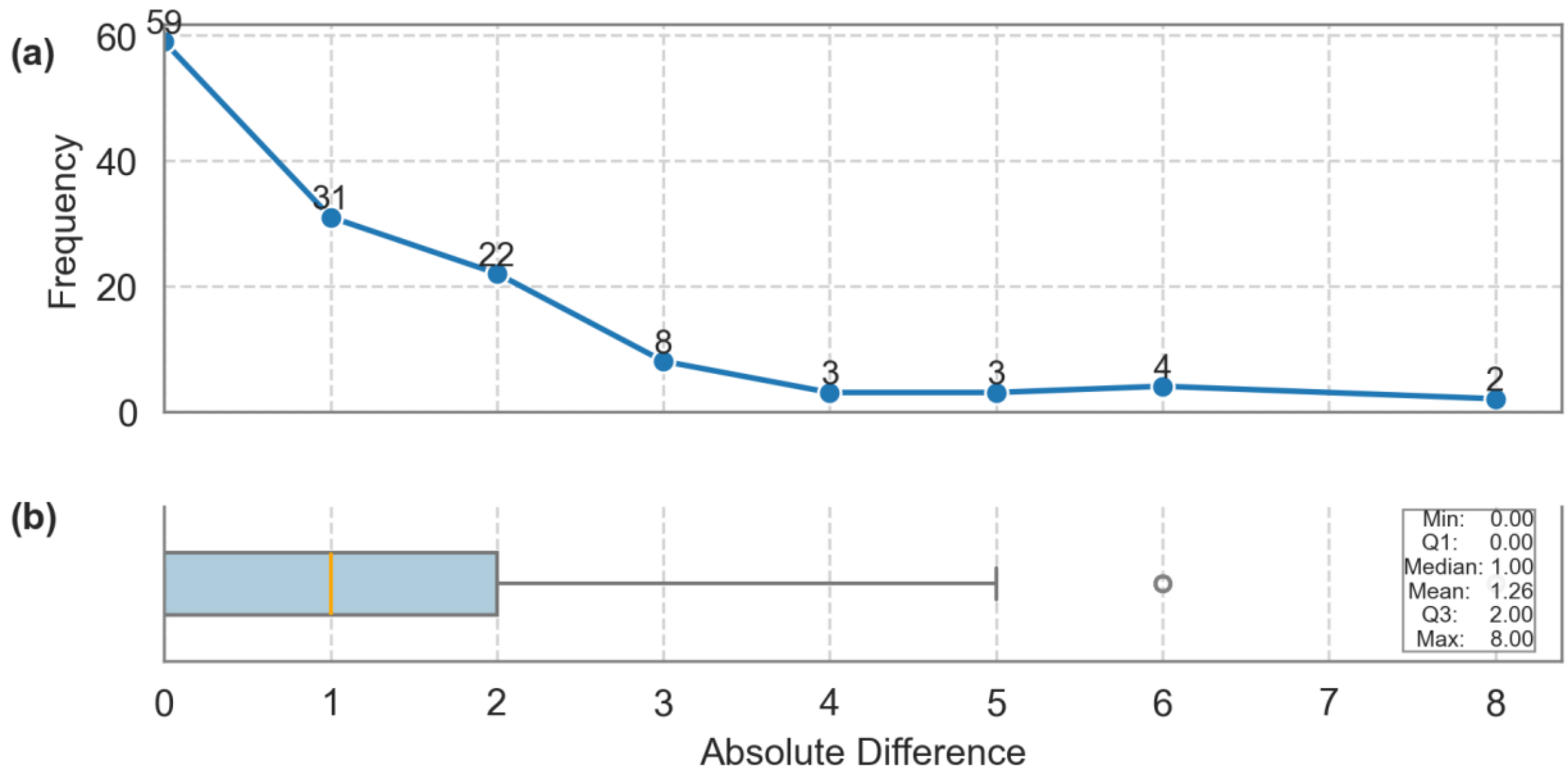

**Fig 3. Distribution and variability of absolute differences between self-reported and HopeBot-assisted PHQ-9 scores (n = 132).**

a, Frequency distribution of absolute differences in PHQ-9 scores between self-administered and HopeBot-assisted assessments.

b, Boxplot summarising the range, central tendency, and outliers of absolute differences between the two formats.

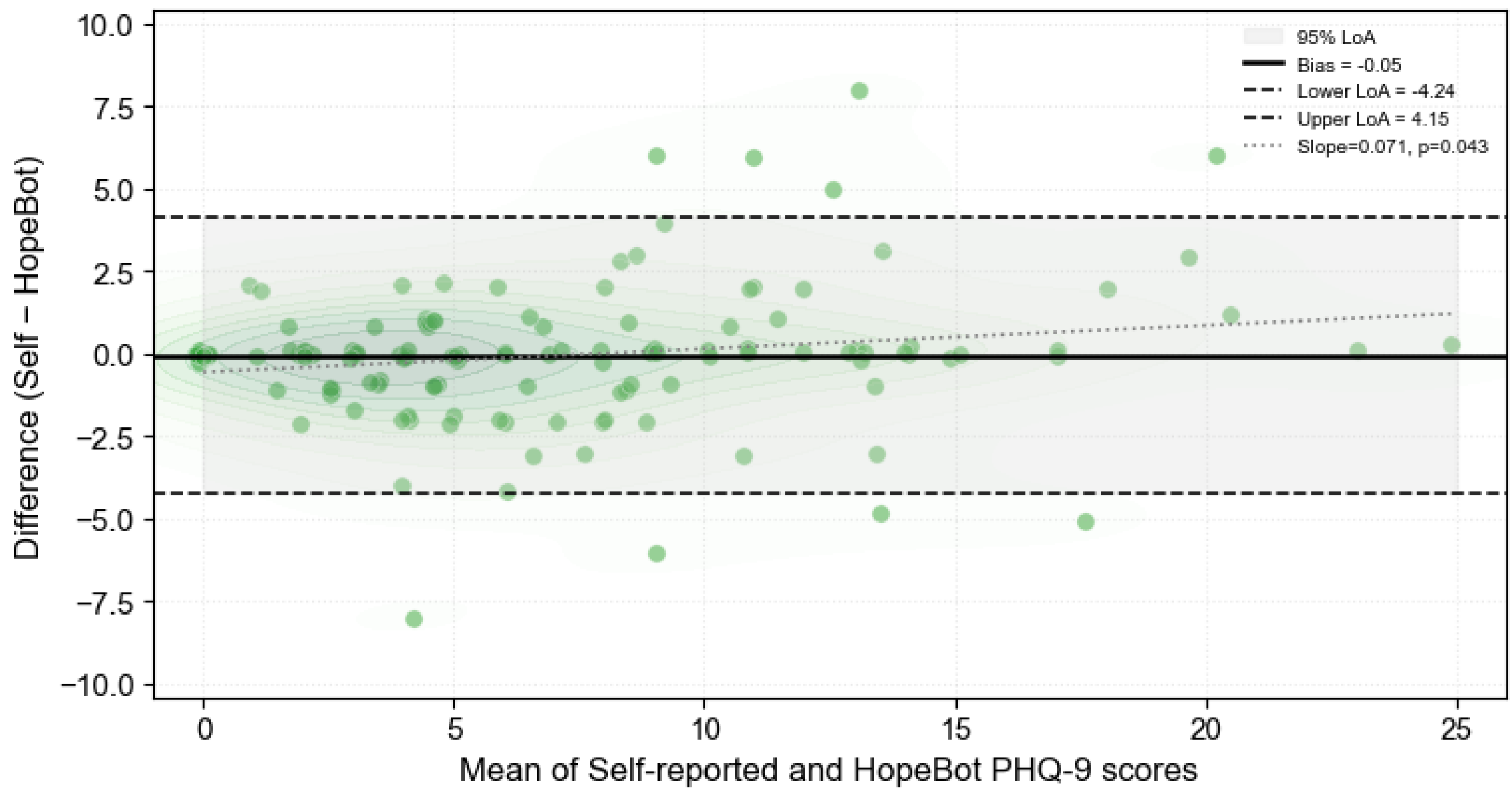

**Fig 4. Bland–Altman plot showing agreement between self-reported and HopeBot-assisted PHQ-9 scores (n**

**= 132).**

Multiple data points overlap due to the discrete 0–27 integer scale of PHQ-9 scoring; slight jitter was applied for visualisation clarity.

# Feedback and User Experience

Participant feedback, coded by mention frequency, highlighted both strengths and limitations of HopeBot. Personalised advice (50 mentions, 17.9%) was the most frequently cited strength, with participants describing the guidance as specific to their situation and clear in next steps. Emotional support (31mentions, 11.1%) was characterised as comforting and non-judgmental, while prompt responses (30 mentions, 10.7%) were valued for efficiency and attentiveness. Criticisms focused on generic or shallow replies (33 mentions, 11.8%) that some users found too broad to act on, and on voice-related issues such as delayed output (10 mentions, 3.6%) or mechanical delivery (8 mentions, 2.9%).

During the PHQ-9 screening phase, 79.5 % of participants described the transition from open dialogue as natural, and 97.7 % found the instructions and questions easy to understand. About one-third (33.3 %, n = 44) requested clarification on item interpretation or scoring; among them, 93.2 % (n = 41/44) found the chatbot's explanations helpful. While 77.3 % characterised the overall interaction as natural, 17 responses (7.0 %) described the transition as abrupt, and 15 (6.1 %) mentioned it felt rushed.

Quantitative ratings reinforced these observations (Fig 5). On a 10-point scale, participants rated HopeBot's handling of sensitive topics mean = 7.60, SD = 1.53, with qualitative comments most frequently citing its empathic tone (63 mentions, 31.0%) and practical guidance (50 mentions, 24.6%). However, several participants also described shallow or generic replies (36 mentions, 17.7%), robotic delivery (16 mentions, 7.9%), and repetitive scripted messages (7 mentions, 3.4%). For example, in response to intense emotional disclosures, HopeBot often reiterated that it was not a licensed psychologist and advised users to seek professional help.

HopeBot's capacity to facilitate emotional expression without judgment received a higher rating (mean = 8.44, SD = 1.53). This was commonly attributed to perceived confidentiality and a non-intrusive communication style. Anonymity was referenced in 72 mentions (38.7%), and 24 mentions (12.9%) highlighted the system's neutral, non-moralising language.

Perceived usefulness of HopeBot's recommendations was moderately high (mean = 7.36, SD = 2.06). Many participants described the advice as clear and actionable (72 mentions, 35.0%), whereas 43 mentions (20.9%) characterised it as overly generic or lacking depth. One participant noted that while the guidance was accurate, its similarity to publicly available information reduced its perceived value.

HopeBot's voice output was rated favourably overall (mean = 7.73, SD = 1.49). Participants frequently highlighted its clear pronunciation (117 mentions, 33.0%) and empathetic, human-like tone (45 mentions, 13.0%), while some noted slow or inaccurate speech recognition (32 mentions, 9.3%) and limited personalisation (25 mentions, 7.2%) as areas for improvement. Despite the positive evaluation of audio quality, nearly half of the participants (45.5%, n = 60) preferred reading the on-screen transcript instead of listening to the full audio, citing greater convenience and discretion. Consistent with this pattern, 51.5% of participants preferred text-based communication, describing it as more practical and private, whereas 40.9% favoured voice interaction for its naturalness and interactivity, and 7.6% expressed no clear preference.

Across all demographic comparisons (Table 2), no statistically significant associations were found between age, gender, ethnicity, education level, recruitment country (UK vs China), or PHQ-9 severity (both self-reported and HopeBot-assisted) and any of the four HopeBot ratings. Employment status produced a significant omnibus effect for the perceived helpfulness of HopeBot's recommendations (Q19: $F = 3.20$, $p = .006$), whereas its impact on the remaining dimensions was nonsignificant (Q17, Q18, Q20: $p > .25$). Follow-up contrasts indicated that the difference reflected variability among employment sub-groups rather than a uniform shift across the full sample.

Participants with prior experience of mental-health treatment gave slightly lower Q19 scores than those without such experience ($t = –2.65$, $p = .012$), whereas ratings for handling sensitive topics (Q17), emotional comfort (Q18), and voice clarity (Q20) did not differ ($p \geq .10$). Previous use of mental-health chatbots was unrelated to any rating (all $p \geq .15$). Feedback patterns were broadly consistent across demographic and cultural subgroups, indicating comparable perceptions of usability and emotional tone.

**Table 2. Association between demographic characteristics and HopeBot user ratings (Q17–Q20).**

| **Demographic variable** | **Test** | **Q17** | | **Q18** | | **Q19** | | **Q20** | |
|---|---|---|---|---|---|---|---|---|---|
| | | **F-statistic** | **p-value** | **F-statistic** | **p-value** | **F-statistic** | **p-value** | **F-statistic** | **p-value** |
| Country of Recruitment (2 levels) | t test | 0.486 | 0.628 | -0.100 | 0.921 | -0.398 | 0.691 | 0.179 | 0.858 |
| Age group (6 levels) | ANOVA | 0.559 | 0.731 | 1.37 | 0.241 | 1.43 | 0.219 | 1.67 | 0.147 |
| Gender (2 levels) | t test | 0.696 | 0.488 | 0.186 | 0.853 | 1.72 | 0.087 | 1.22 | 0.227 |
| Ethnicity (5 levels) | ANOVA | 0.889 | 0.473 | 0.398 | 0.810 | 0.821 | 0.514 | 0.709 | 0.587 |
| Highest education (4 levels) | ANOVA | 0.174 | 0.951 | 0.831 | 0.508 | 1.18 | 0.323 | 0.607 | 0.658 |
| Employment status (6 levels) | ANOVA | 0.661 | 0.681 | 1.32 | 0.253 | 3.20 | **0.006** | 1.18 | 0.320 |
| Mental health support experience (binary) | t test | -0.637 | 0.528 | -1.53 | 0.137 | -2.65 | **0.012** | -1.14 | 0.261 |
| Chatbot experience (binary) | t test | -1.44 | 0.153 | 0.380 | 0.705 | -0.859 | 0.392 | 0.134 | 0.894 |
| PHQ-9 Severity Result – HopeBot (5 levels) | ANOVA | 1.23 | 0.300 | 0.462 | 0.763 | 1.57 | 0.188 | 1.72 | 0.150 |
| PHQ-9 Severity Result – Self-test (5 levels) | ANOVA | 1.40 | 0.238 | 0.284 | 0.888 | 0.823 | 0.513 | 0.394 | 0.813 |

Notes. Q17 = handling of sensitive depression topics; Q18 = comfort expressing feelings without judgment; Q19 = helpfulness of recommendations; Q20 = clarity and tone of voice output. All values are rounded to three significant figures. Bold indicates p < 0.05.

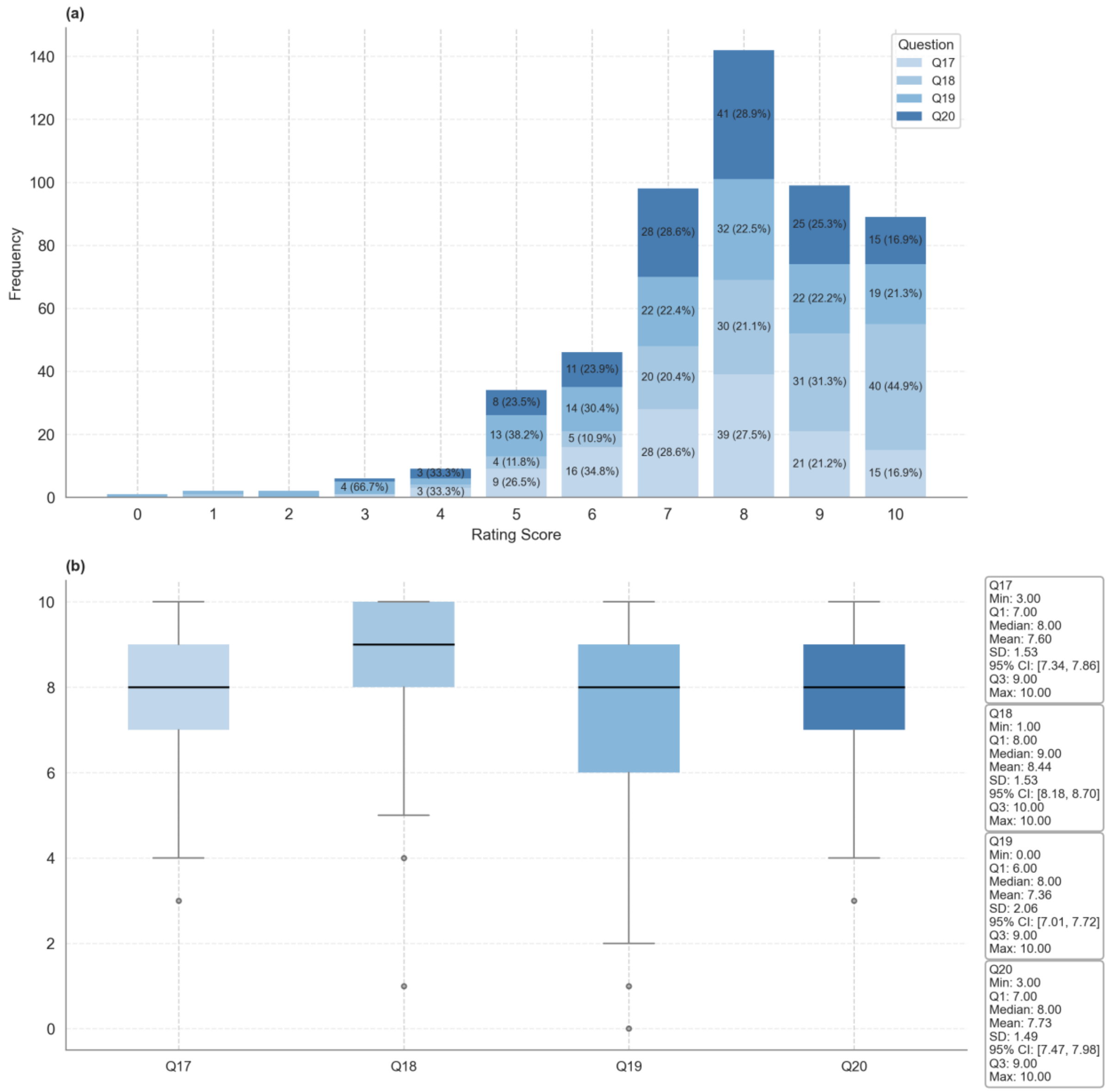

**Fig 5. Distribution histograms and boxplots for four HopeBot evaluation items (Q17–Q20).**

a, Stacked bar plot illustrating the frequency distribution of user ratings (0–10) across four evaluation items. Each bar represents the total number of responses at each score point, subdivided by item. Higher scores indicate more positive user evaluations. Numeric labels within each segment indicate absolute counts and corresponding percentages. To maintain legibility and avoid overcrowding, only segments with sufficient height (e.g., ≥2 units) display numeric labels. Low-frequency responses (e.g., ratings 0–4) are fully retained in the underlying analysis but may not be annotated if their bar height falls below the display threshold.
b, Boxplots summarising the score distributions for each evaluation item, with accompanying descriptive statistics including minimum, maximum, IQR, median, mean, SD, and 95% CI. These values are displayed to the right of each box for easy comparison.
Notes. Q17 = handling of sensitive depression topics; Q18 = comfort expressing feelings without judgment; Q19 = helpfulness of recommendations; Q20 = clarity and tone of voice output.

## Perceived Acceptability and Adoption Intentions

Participant preferences for PHQ-9 administration formats varied. A majority (n = 92, 69.7%) indicated a preference for the chatbot-assisted format over self-completion, citing that it felt more engaging (39 mentions, 20.2%), emotionally supportive and interactive (36 mentions, 18.7%), and provided real-time clarification and guidance during question interpretation (32 mentions, 16.6%). In contrast, participants who preferred self-administration emphasised its efficiency (20 mentions, 10.4%) and its suitability for situations without immediate emotional distress (13 mentions, 6.7%). A small subset (n = 6, 4.6%) expressed no clear preference. Preference was associated with employment status overall ($\chi^2 = 21.69$, df = 12, p = .041), but no individual subgroup (all p > .5) showed a statistically significant odds ratio, suggesting weak or diffuse effects.

Among all respondents, only 19.7% (n = 26) reported prior engagement with professional mental-health services, yet all participants were invited to reflect on HopeBot's perceived performance relative to human counselling. Consistent with earlier themes, participants positively appraised the chatbot's structured questioning and empathic tone (each 38 mentions, 12.3%). However, limitations were frequently noted, including perceptions of insufficient human-likeness (58 mentions, 18.7%), emotional shallowness or detachment (42 mentions, 13.5%), and overly generic or impersonal responses lacking individual tailoring (19 mentions, 6.1%).

Regarding adoption, most participants (n = 115, 87.1%) indicated willingness to use or recommend HopeBot in the future. Many perceived its potential to support mental health screening, particularly through early detection and emotionally responsive communication (56 mentions, 22.0%; 32 mentions, 12.6%). Other cited advantages included immediate accessibility (15 mentions, 5.9%), rapid response time (14 mentions, 5.5%), and anonymous interaction (8 mentions, 3.1%). Several respondents (n = 12, 4.7%) emphasised that such tools should complement rather than replace professional care. Concerns focused on the need for clinical validation (4 mentions, 1.6%), possible diagnostic unreliability (3 mentions, 1.2%), and data privacy risks (11 mentions, 4.4%). Nonetheless, willingness to recommend was significantly lower among participants with prior mental health service experience (Fisher's exact OR = 0.22, CI 0.05–0.92, p = .0497).

# Discussion

## Principal Findings

This study demonstrates the feasibility of administering the PHQ-9 through a GPT-4o-powered, voice-interactive chatbot (HopeBot). HopeBot-assisted and self-administered scores showed high concordance (ICC = 0.92; median absolute difference = 1 point), supporting its feasibility as a reliable administration method rather than implying diagnostic equivalence. Participants generally found the chatbot easy to use and responsive, consistent with usability ratings and suggesting that conversational delivery can improve engagement within standardised screening frameworks.

Although clinician-administered PHQ-9 interviews detect suicidality and comorbid conditions with higher sensitivity [15], self-administered formats remain standard in digital screening, showing acceptable psychometric performance (sensitivity ≈0.80; specificity ≈0.85) [11,54]. The self-report version was therefore used as a pragmatic comparator, reflecting real-world contexts in which individuals typically complete questionnaires independently. HopeBot's strong concordance with self-report supports the fidelity of LLM-assisted administration, but further validation against clinician-led interviews is required to determine diagnostic accuracy and clinical applicability.

Beyond score concordance, HopeBot elicited considerable user trust. Among the 75 participants who directly compared the two formats, 70.7% (n = 53/73) expressed greater confidence in the chatbot-assisted scores, attributing this preference to features such as real-time clarification (33 mentions, 42.9%) and an empathic tone (15 mentions, 19.5%). These interactional features partially resembled elements of semi-structured interviews, enhancing engagement and clarity while preserving the scalability and standardisation of automated delivery.

Perceptions of recommendation helpfulness (Q19) varied across subgroups. Full-time students and individuals managing household responsibilities gave lower ratings than full-time employees ($F = 3.20$, $p = .006$), and those with prior mental health service experience rated recommendations less helpful than first-time users ($t = -2.65$, $p = .012$). These differences likely reflect heightened expectations shaped by users' life context and therapeutic background. Students and homemakers—often managing complex emotional demands with limited external support—may have anticipated greater empathy and personalisation [55,56]. Similarly, individuals with prior counselling experience may have evaluated responses against professional standards [7], consistent with expectancy-disconfirmation theory [57]. These findings suggest that trust and perceived usefulness of AI-based mental-health tools are shaped by users' experiential backgrounds, underscoring the need for adaptive designs that accommodate diverse expectations of empathy and personalisation.

## Comparison with Prior Systems

As summarised in Table 3, HopeBot introduces several substantive advancements over earlier PHQ-9 chatbots that rely primarily on Dialogflow-based intent matching (e.g., Perla [18],

Marcus [27], DEPRA [21]). By integrating RAG with the GPT-4o architecture, HopeBot supports fully open-ended interaction while meeting the technical constraints of real-time screening. GPT-4o was selected based on three key considerations: (1) independent benchmarks reported the lowest latency among publicly available LLMs (≈0.45 s for text, ≈0.32 s for audio) at that time, outperforming Claude 3 and Gemini 1.5 [59];(2) its unified multimodal framework eliminates the need for separate ASR–TTS pipelines, which remain necessary for open-source and contemporary commercial alternatives [60]; and (3) its extended context window (100k tokens) and multilingual tokeniser ensure compatibility with the demands of interactive PHQ-9 delivery while preserving clinical safety constraints [61]. Although LLMs such as Claude 3, Gemini 1.5, and Llama-3 warrant future evaluation, their higher latency and limited support for speech and alignment control made them unsuitable for the present implementation.

This architecture enables dynamic turn-taking, clarification of ambiguous responses, and seamless support for languages beyond English and Mandarin—capabilities not reported in prior systems. Transparency is further enhanced through item-level scoring and source-linked rationales, features absent in comparators such as IGOR [26] and Marcus [27]. Among participants who engaged with the clarification module, 93.2% indicated that these explanations improved their response accuracy.

While HopeBot was not designed to substitute for clinician-led diagnosis, its findings indicate potential utility as a preliminary engagement or triage aid within mental-health screening workflows [62]. Unlike earlier rule-based systems that restricted user expression [63], HopeBot's adaptive clarification maintains alignment with the PHQ-9 item structure while allowing users to articulate symptoms more naturally, which may enhance clarity and perceived relevance.

In practical terms, this standardised yet conversational approach may support early identification of depressive features, particularly among individuals who might otherwise defer or avoid assessment. The high concordance with self-administered PHQ-9 scores (ICC = 0.92) suggests that conversational delivery can preserve quantitative integrity while improving accessibility and engagement. Collectively, these characteristics suggest that HopeBot could function as a research-oriented adjunct bridging self-report and clinician-administered formats, providing a framework for scalable and low-risk screening.

User feedback provides further support for these procedural benefits. In contrast to Marcus, where only 18.1% of users preferred the chatbot over conventional self-report [27], 69.7% (n = 92) of participants in this study favoured HopeBot, and 87.1% (n = 115) expressed willingness to reuse or recommend it. These results indicate higher levels of user acceptance and engagement than those reported for earlier rule-based systems. Combined with its low-latency performance, bilingual adaptability, and transparent item-level explanations, this evidence provides preliminary, exploratory support for HopeBot's feasibility and user alignment within future digital screening research workflows.

## Table 3. Comparison of automated depression screening tools across key functional dimensions

| Dimension | Core architecture | Language flexibility | Screening instrument | Explainability | User study (N) | Empathy/tone | Deployment potential | Key limitation in prior work |
|---|---|---|---|---|---|---|---|---|
| **Perla (2020) [18]** | Google Dialogflow with ML-based intent classification and Firebase backend | Natural language input with ~200 phrases per item and synonym matching | PHQ-9 (Spanish) | Provides total score, risk status, and resource links at the end | 276 participants; 108 completed both Perla and web-based PHQ-9 | Supportive tone with female persona and encouraging prompts | Web and major messaging platforms (e.g., Messenger, Google Assistant, Telegram) | Limited validation, English-only tools, and low engagement in prior form-based tools |
| **Marcus (2023) [27]** | Dialogflow intents + BERT model (Node.js / GCP; Kommunicate UI) | Free-text input classified to PHQ-9 | PHQ-9 (English) | Outputs total PHQ-9 score only | 81 U.S. college students (130 enrolled) | Neutral; static male avatar; no empathy modelling | iOS app and web chat prototype | Earlier, PHQ-9 chatbots lacked validation with U.S. college students and used only fixed-choice input |
| **IGOR (2021) [26]** | Dialogflow intents with Node.js + Firebase backend | Button/option input (PHQ-9 scores 0–3) | PHQ-9 | Sends total score to clinician; not shown to user | 10 university staff (usability test) | Neutral; no empathic responses | Prototype within MS self-management app | Results hidden from the user; rule-based flow fails on off-topic input |
| **DEPRA (2023) [21]** | Dialogflow chatbot with 27 SIGH-D/IDS-C intents | Open-text input with intent matching | SIGH-D + IDS-C | Final score and severity level only; no item-level feedback | 50 Australian adults | Neutral tone; no empathy modelling | Facebook Messenger chatbot prototype | High cognitive load; long completion time |
| **EmoScan (2024) [28]** | Mistral-7B fine-tuned on synthetic clinical interviews (PsyInterview) | Free-text, multi-turn inputs; fine-tuned LLM | DSM-5-based emotional disorder classification (coarse & fine-grained) | LLM-generated explanations based on DSM-5 criteria | 1,157 synthetic cases; 50 expert-evaluated; GPT-4-based performance evaluation | Simulated empathy assessed by GPT-4 and clinical experts | Research prototype; not deployed clinically | Heavily synthetic Reliance on synthetic data; limited real-world generalisability |
| **Moodpath App (2021) [58]** | Smartphone app, 3× daily AA (45 ICD-10 items + mood) | Tap yes/no → 4-level severity; 5-point mood scale | DSM-5-based emotional disorder classification (coarse & fine-grained) | 14-day summary with score, severity band & mood charts | 113 general-population users | Neutral; no empathic dialogue | Live on iOS & Android | Prior tools used retrospective questionnaires with little validation |
| **HopeBot (2025)** | GPT-4o with RAG | Supports open-ended user input in various languages | PHQ-9 administered via free-text/voice dialogue, with dynamic clarification and fuzzy score interpretation | Provides item-level rationales and context-relevant evidence drawn from curated knowledge bases | 132 participants (Chinese + UK residents) | LLM-generated responses were generally perceived as empathic; mean comfort rating 8.51/10, though variation in emotional tone and delivery style was reported | Available via web interface; compatible with both desktop and mobile devices (iOS, Android); supports both Mandarin and English voice/text modalities | Absence of non-verbal cues, occasional mechanical tone in voice output, and lack of clinical validation for diagnostic reliability |

## Technical, Ethical, and Linguistic Constraints

Despite its technical strengths, HopeBot did not fully replicate the relational depth of professional counselling. A substantial number of participants described the interaction as emotionally flat or impersonal (58 mentions, 18.7%), citing insufficient affective nuance (42 mentions, 13.5%) and reliance on generic responses (19 mentions, 6.1%). These limitations indicate that, even with prompt tuning and retrieval augmentation, simulated empathy remains perceptibly artificial. These limitations underscore the persistent gap between linguistic fluency and emotional authenticity in LLM-mediated dialogue.

More broadly, these limitations reflect structural constraints inherent in current-generation conversational AI [64]. While current LLMs can generate contextually coherent text, they lack access to non-verbal cues such as tone, expression, or posture that clinicians rely on to assess distress [22]. Cross-linguistic speech synthesis poses further challenges. Although GPT-4o natively supports Mandarin text, the deployment of a text-to-speech engine optimised for English prosody introduced prosodic inconsistencies that reduced perceived naturalness and constrained affective expressiveness [65]. Future developments in culturally adaptive voice models may improve emotional fidelity across languages.

Beyond technical issues, ethical and epistemic considerations remain. Repeated exposure to polished AI language may subtly influence how users articulate emotional experiences [66]. Clinicians may also adjust judgments when confronted with opaque AI recommendations, even in the absence of a clear clinical rationale [67], raising concerns about automation bias and the erosion of clinical autonomy. Accordingly, systems such as HopeBot should be positioned as adjunctive tools that support, rather than replace, professional expertise [22]. Responsible deployment will require transparent algorithmic logic, explicit clinical oversight, and safeguards that ensure interpretability, accountability, and informed consent. Future research should extend beyond screening accuracy to investigate how such technologies influence therapeutic relationships, user trust, and long-term mental health outcomes.

## Study Limitations

Several methodological and contextual limitations warrant acknowledgement. First, the validation framework compared HopeBot-assisted and self-administered formats rather than clinician-led references; results thus reflect format-level concordance rather than diagnostic validity. Second, the sequential completion of both PHQ-9 versions within one session may have introduced recall bias; counterbalanced or time-separated designs should be explored. Third, the sample skewed younger and digitally literate, which may overestimate usability in older or digitally excluded groups. Fourth, while both English and Mandarin participants were included, subgroup trends were reported in the Results section; the present study was not intended to test cultural differences. Fifth, while both quantitative and qualitative subgroup comparisons were conducted, some demographic strata (e.g., ethnic minority or severe-symptom groups) were small, limiting the statistical power for detecting fine-grained effects. In addition, the controlled testing environment may not reflect naturalistic conditions. All findings are specific to GPT-4o and may not extend to other LLM-based systems.

## Future Research

Future research should evaluate clinical utility, safety, and equity across health-care settings. Multisite and randomised trials comparing LLM-assisted, self-administered, and clinician-administered PHQ-9 formats would clarify diagnostic performance and workflow impact. Development of governance frameworks, including escalation protocols, audit trails, and transparent disclosures, will be essential for regulatory compliance. Particular attention is needed for high-risk encounters requiring non-verbal cue interpretation and for underserved populations with limited digital access. Cross-cultural extensions, already preliminarily supported in this cohort, should be validated across health systems in both the UK and China.

## Conclusion

HopeBot feasibly administers the PHQ-9 with high scoring fidelity and user acceptability. By integrating conversational empathy with validated screening logic, it may serve as a scalable adjunct for early depression detection. Continued evaluation across clinical and cultural contexts will be essential to ensure responsible and equitable translation into practice.

# Acknowledgements

The authors gratefully acknowledge the invaluable contributions of the following colleagues to the evaluation of HopeBot: Dr Alexandru Petcu, MD (NHS consultant psychiatrist) for clinical insights and safety guidance; Kai Yao and Zuyu Wang (Post-Graduate Teaching Assistants, UCL Division of Psychiatry) for assistance with study design, participant recruitment and data interpretation; and Wei'an Li (licensed mental-health counsellor, China) for expert feedback on Mandarin content and cultural adaptation. Their support greatly strengthened the rigour and relevance of this work.

# Data availability

Restrictions apply to the availability of the full dataset generated and analysed during the current study in order to protect participant privacy; accordingly, these data are not publicly available. However, the custom training data used to develop the RAG component of HopeBot is available at: https://github.com/candiceguo0528/HopeBot-Candice.

# Code availability

The code for the analysis is available through a GitHub code repository (https://github.com/candiceguo0528/HopeBot-Candice).

# Funding

This research received no specific grant from any funding agency in the public, commercial, or not-for-profit sectors.

# Competing Interests

The authors declare that they have no competing interests.

# Supporting information

## S1 Table. GRRAS-checklist.

| Section | Item # | Checklist item | Reported on page # |
|---|---|---|---|
| Title/Abstract | 1 | Identify in title or abstract that interrater/intrarater reliability or agreement was investigated. | 1-2 |
| Introduction | 2 | Name and describe the diagnostic or measurement device of interest explicitly. | 3-4 |
| | 3 | Specify the subject population of interest. | 3-4 |
| | 4 | Specify the rater population of interest (if applicable). | 3-4 |
| | 5 | Describe what is already known about reliability and agreement and provide a rationale for the study (if applicable). | 4-5 |
| Methods | 6 | Explain how the sample size was chosen. State the determined number of raters, subjects/objects, and replicate observations. | 9 |
| | 7 | Describe the sampling method. | 9 |
| | 8 | Describe the measurement/rating process (e.g. time interval between repeated measurements, availability of clinical information, blinding). | 8-9 |
| | 9 | State whether measurements/ratings were conducted independently. | 9 |
| | 10 | Describe the statistical analysis. | 9-10 |
| Results | 11 | State the actual number of raters and subjects/objects which were included and the number of replicate observations which were conducted. | 9&11 |
| | 12 | Describe the sample characteristics of raters and subjects (e.g. training, experience). | 10-12 |
| | 13 | Report estimates of reliability and agreement including measures of statistical uncertainty. | 12-14 |
| Discussion | 14 | Discuss the practical relevance of results. | 19-20 |
| Auxiliary material | 15 | Provide detailed results if possible (e.g. online). | 27 |

**S2 Table. STARD2015 Checklist.**

| Section & Topic | No | Item | Reported on page # |
|---|---|---|---|
| **TITLE OR ABSTRACT** | | | |
| | 1 | Identification as a study of diagnostic accuracy using at least one measure of accuracy (such as sensitivity, specificity, predictive values, or AUC) | |
| **ABSTRACT** | | | |
| | 2 | Structured summary of study design, methods, results, and conclusions (for specific guidance, see STARD for Abstracts) | 1-2 |
| **INTRODUCTION** | | | |
| | 3 | Scientific and clinical background, including the intended use and clinical role of the index test | 3-4 |
| | 4 | Study objectives and hypotheses | 4-5 |
| **METHODS** | | | |
| **Study design** | 5 | Whether data collection was planned before the index test and reference standard were performed (prospective study) or after (retrospective study) | 8&9 |
| **Participants** | 6 | Eligibility criteria | 9 |
| | 7 | On what basis potentially eligible participants were identified (such as symptoms, results from previous tests, inclusion in registry) | |
| | 8 | Where and when potentially eligible participants were identified (setting, location and dates) | 9 |
| | 9 | Whether participants formed a consecutive, random or convenience series | 9 |
| **Test methods** | 10a | Index test, in sufficient detail to allow replication | 5 |
| | 10b | Reference standard, in sufficient detail to allow replication | 5 |
| | 11 | Rationale for choosing the reference standard (if alternatives exist) | 5 |
| | 12a | Definition of and rationale for test positivity cut-offs or result categories of the index test, distinguishing pre-specified from exploratory | |
| | 12b | Definition of and rationale for test positivity cut-offs or result categories of the reference standard, distinguishing pre-specified from exploratory | |
| | 13a | Whether clinical information and reference standard results were available to the performers/readers of the index test | |
| | 13b | Whether clinical information and index test results were available to the assessors of the reference standard | |
| **Analysis** | 14 | Methods for estimating or comparing measures of diagnostic accuracy | 9-10 |
| | 15 | How indeterminate index test or reference standard results were handled | |
| | 16 | How missing data on the index test and reference standard were handled | 10 |
| | 17 | Any analyses of variability in diagnostic accuracy, distinguishing pre-specified from exploratory | 9-10 |
| | 18 | Intended sample size and how it was determined | 9 |
| **RESULTS** | | | |
| **Participants** | 19 | Flow of participants, using a diagram | |
| | 20 | Baseline demographic and clinical characteristics of participants | 10-12 |
| | 21a | Distribution of severity of disease in those with the target condition | 12 |
| | 21b | Distribution of alternative diagnoses in those without the target condition | 12 |
| | 22 | Time interval and any clinical interventions between index test and reference standard | |
| **Test results** | 23 | Cross tabulation of the index test results (or their distribution) by the results of the reference standard | 12-13 |
| | 24 | Estimates of diagnostic accuracy and their precision (such as 95% confidence intervals) | 12-13 |
| | 25 | Any adverse events from performing the index test or the reference standard | |
| **DISCUSSION** | | | |
| | 26 | Study limitations, including sources of potential bias, statistical uncertainty, and generalisability | 22 |
| | 27 | Implications for practice, including the intended use and clinical role of the index test | 20 |
| **OTHER INFORMATION** | | | |
| | 28 | Registration number and name of registry | 5 |
| | 29 | Where the full study protocol can be accessed | 5 |
| | 30 | Sources of funding and other support; role of funders | 23 |

**S3 Text. Ethics, safety, and privacy.**

HopeBot was explicitly framed as a supportive screening tool rather than a diagnostic or emergency service and was designed to communicate in a non-judgmental and empathetic manner.

The system used predefined keyword triggers to detect psychological crises, such as references to self-harm or acute emotional distress. When activated, it interrupted the session, delivered a pre-scripted supportive message, and redirected users to appropriate mental health services. A curated list of country-specific resources was integrated: for UK users, this included Samaritans (116 123) and Shout (text 85258); for users in China, it included major psychiatric institutions such as Peking University Sixth Hospital and Shanghai Mental Health Centre. All safety protocols and referral procedures were approved by the UCL Research Ethics Committee under a high-risk ethics application.

To ensure privacy and data security, all interactions were anonymised and stored in compliance with the General Data Protection Regulation. No personally identifiable data was collected, and OpenAI's API settings were configured to prevent data retention. Data access was restricted to authorised research personnel and subject to participant consent.

**S4 Text. 25-item questionnaire.**

1. What is your age?
2. With which gender do you identify most?
3. What is your ethnic background?
4. What is your level of education?
5. What is your employment status?
6. Can you tell me a little bit about your background, including any experience you've had with mental health support or chatbots?
7. How familiar are you with large language models, such as ChatGPT, Gemini, Bing or similar AI technologies?
8. During your interaction with HopeBot, was there anything that stood out to you—either positively or negatively? Could you describe a specific moment or feature that made an impression on you and explain why?
9. Were there moments during your conversations with HopeBot when you felt truly understood and heard? Were there times when its responses felt a bit off or robotic? How did these experiences impact you?
10. What is your Patient Health Questionnaire-9 (PHQ-9) self-reported result?
11. What is your PHQ-9 result, as assessed with the assistance of HopeBot?
12. Is there a difference between the results of your PHQ-9 self-test and the scores completed with the assistance of HopeBot? If so, what do you think might have caused this discrepancy? Which results do you find more trustworthy, and why?
13. How does using HopeBot to complete the PHQ-9 compare to completing it alone at home? Which did you prefer, and why?
14. Did HopeBot introduce the PHQ-9 in a natural way? Were there any moments when its introduction felt disruptive or interrupted your conversation? Please share your thoughts.
15. Were the PHQ-9 questions easy to understand and answer? If anything was unclear, did you try asking HopeBot for help? And while you were responding, did HopeBot offer any useful support?
16. Did your conversation with HopeBot feel natural? Were there any points where the flow felt forced or responses seemed robotic?
17. On a scale of 1-10, how well did HopeBot handle sensitive topics related to depression? What influenced your rating?
18. On a scale of 1-10, how comfortable did you feel expressing your feelings without judgment? What influenced your rating?
19. On a scale of 1-10, how helpful were HopeBot’s recommendations? What influenced your rating?
20. On a scale of 1 to 10, how clear was HopeBot’s voice output? Did its tone match your expectations? Were you willing to listen to the entire response? Please provide the reasoning behind your rating.
21. Do you prefer using voice or text when chatting with HopeBot? Why? Which one feels more natural, like talking to a real person, and which is more convenient for you?
22. In your experience using HopeBot, in what aspects do you believe its mental health support services approach the level of professional mental health practitioners?

Conversely, in which areas do you find it still difficult to serve as a substitute? Please elaborate on your perspective.

23. What were HopeBot’s strengths and weaknesses, and how could it better support mental health and depression screening?
24. Would you feel comfortable using HopeBot in the future or suggesting it to someone else? What factors influence your decision?
25. How do you see AI and technology shaping the future of mental health screening and support?

**S5 Table.PHQ-9 score differences (HopeBot – Self-administered) across demographic and experience variables.**

| Variable | Group | Median | IQR | Test | Statistic | p_raw | p_holm |
|---|---|---|---|---|---|---|---|
| **Age group (years)** | 18–24 years old | 0.00 | 2.00 | Kruskal–Wallis H | 6.565 | 0.766 | 1.000 |
| | 25–34 years old | 0.00 | 0.25 | | | | |
| | 35–44 years old | 0.00 | 0.75 | | | | |
| | 45–54 years old | 0.00 | 3.50 | | | | |
| | 55–64 years old | 0.00 | 1.00 | | | | |
| | 65–70 years old | 0.00 | 2.00 | | | | |
| **Gender** | Female | 0.00 | 1.00 | Mann–Whitney U | 2243.5 | 0.690 | 1.000 |
| | Male | 0.00 | 2.00 | | | | |
| **Ethnicity** | Asian or Asian British | 0.00 | 2.00 | Kruskal–Wallis H | 4.646 | 0.326 | 1.000 |
| | Black / Black British / Caribbean | 0.50 | 2.00 | | | | |
| | Mixed or Multiple groups | 0.50 | 0.50 | | | | |
| | White | 0.00 | 0.00 | | | | |
| | Prefer not to say | 3.00 | 0.00 | | | | |
| **Highest education** | No formal | 0.00 | 0.00 | Kruskal–Wallis H | 2.140 | 0.952 | 1.000 |
| | Further education | 0.50 | 1.25 | | | | |
| | Undergraduate degree | 0.00 | 1.25 | | | | |
| | Postgraduate degree | 0.00 | 0.00 | | | | |
| | Prefer not to say | 0.00 | 0.00 | | | | |
| **Employment status** | Full-time employment | 0.00 | 1.00 | Kruskal–Wallis H | 8.130 | 0.616 | 1.000 |
| | Part-time employment | 0.00 | 0.50 | | | | |
| | Full-time education / training | 0.00 | 0.00 | | | | |
| | Looking after the home | 0.00 | 0.00 | | | | |
| | Retired | 1.00 | 1.00 | | | | |
| | Home | −2.00 | 1.50 | | | | |
| | Other | 0.00 | 0.50 | | | | |
| | Prefer not to say | −1.00 | 0.00 | | | | |

| Variable | Group | Median | IQR | Test | Statistic | p_raw | p_holm |
|---|---|---|---|---|---|---|---|
| | Student | 0.00 | 3.00 | | | | |
| **Familiarity with LLMs** | HEARD_ONLY | 1.00 | 1.00 | Kruskal–Wallis H | 13.313 | 0.021 | 0.165 |
| | NO_EXPERIENCE | −3.00 | 3.00 | | | | |
| | OCCASIONAL_USER | −1.00 | 2.25 | | | | |
| | REGULAR_USER | 0.00 | 1.00 | | | | |
| | TECHNICAL_EXPERT | 0.00 | 0.50–7.00 | | | | |
| **Mental health chatbot experience** | No | 0.00 | 1.00 | Mann–Whitney U | 1818.0 | 0.646 | 1.000 |
| | Yes | 0.00 | 2.00 | | | | |
| **Previous mental health support experience** | No | 0.00 | 2.00 | Mann–Whitney U | 1112.0 | 0.111 | 0.774 |
| | Yes | 0.00 | 1.75 | | | | |

**S5 Text. Subgroup analysis of PHQ-9 score differences between HopeBot-assisted and self-administered assessments.**

Group-wise median and interquartile range (IQR) of PHQ-9 score differences (HopeBot – Self-administered) by demographic and experience variables.
For each subgroup, medians and IQRs are presented together with the corresponding nonparametric test statistics. Mann–Whitney U tests were used for binary variables and Kruskal–Wallis H tests for multi-category variables. P values were corrected for multiple comparisons using the Holm method.

No subgroup differences reached statistical significance after correction (all $p > 0.05$), suggesting that the differences between HopeBot-assisted and self-administered PHQ-9 scores were stable across age, gender, ethnicity, education, employment, familiarity with large language models (LLMs), and prior mental-health-related experiences.

**S6 Text. Codebook.**

This supplementary material presents the complete qualitative codebook developed for thematic analysis of user experiences with HopeBot. The qualitative analysis covered a total of 18 core qualitative questions (Q7–Q9, Q12–Q14, Q16-Q20, Q22–Q25) and provided coding guidance for additional optional questions.

Q7 Familiarity with large language models
Theme 1 Technical Familiarity — Codes: NO_EXPERIENCE, HEARD_ONLY, OCCASIONAL_USER, REGULAR_USER, TECHNICAL_EXPERT

Q8 Memorable moments during HopeBot use
Theme 1 User Experience & Usability — Codes: Alt-tool preference; Desire for more interaction; Dislike text-reading; Easy to use; General positive impression; No special impression; Novel & unique experience; Skill-dependent usability
Theme 2 Human-likeness & Realism — Codes: Feeling cared for; Genuine & meaningful interaction; Human-like interaction
Theme 3 Voice Quality & Audio — Codes: Desire for voice-output control; Positive voice experience; Unnatural voice; Voice annoyance / unnecessary output; Voice overlap / disruption; Voice timing issue
Theme 4 Interaction Flow & Responsiveness — Codes: Direct questions; Disrupted conversation; Lack depth / analysis; Lack personalisation; Quick & instant response; Responsive interaction; Slow response / delay; Smooth & engaging conversation
Theme 5 Emotional Support & Empathy — Codes: Affirmation / validating feelings; Emotional support; Encouragement; Guidance toward positive reflection / action; Response to emotional distress
Theme 6 Trust & Privacy — Codes: Disclosure control; Feeling analysed; Privacy concern
Theme 7 Modality Preference — Codes: Typing preference; Voice-to-text preference
Theme 8 Helpfulness & Content — Codes: Adequate amount of advice; Generic / inadequate response; Specific & personalised advice; Targeted health & sleep advice; Timely & relevant suggestions
Theme 9 Technical & Accuracy — Codes: Accurate voice recognition; System functionality issues; Voice recognition problems

Q9 Feeling understood versus robotic
Theme 1 Feeling Understood & Heard —Codes: Affirmation & validation; Contextual response; Deep empathy; Emotional support; Empathetic questioning; Feeling cared for/understood; Genuine & meaningful interaction; Human-like interaction; Encouraged sharing/increased willingness to share
Theme 2 Feeling "Off" or Robotic — Codes: Generic/ inadequate/ formulaic/ repetitive replies; Lack of personalisation; Lack of depth/ analysis; No emotional connection; Stiff or monotonous tone; Mechanical/ unnatural voice delivery; Desire for greater voice-output control

Theme 3 Emotional/ Psychological Impact — Codes: Positive emotional impact & stress relief;

Neutral/ mixed emotional impact; Negative emotional impact; Emotional encouragement; Guidance toward positive reflection/ action

Theme 4 Interaction & Trust Dynamics — Codes: Quick & instant response; Slow response/ delay; Smooth & engaging conversation; Increased sharing/ openness; Reduced engagement/ willingness to engage; Trust shift/ no attachment

Theme 5 Perceived Quality & Expectations — Codes: Overall satisfaction/ general positive impression; Specific & personalised advice; Timely & relevant suggestions; Targeted health & sleep advice; Need for more human-like features; Tool comparison/ perception of professionalism; Clinical perception; Requests for specific improvements

Q12 Difference between PHQ-9 scores
Theme 1 Reasons for Difference — Codes: R-DEEPER, S-TIME, S-VOICE_ERR, S-TECH_LIMIT, R-EMO_SUP, R-GUIDE, S-PRIVACY, S-MOOD, S-INTUIT

Q13 HopeBot versus self-assessment
Theme 1 Process Efficiency — Codes: P-FAST, P-SLOW, P-EASY
Theme 2 Interaction Experience — Codes: I-ENGAGE, I-HUMAN, I-NONJUDGE
Theme 3 Cognitive & Result Quality — Codes: C-GUIDE, C-FEEDBK, C-CONTEXT, C-OBJECT, C-INTUIT
Theme 4 Risk & Concerns — Codes: R-PRIVACY, R-TECH, R-FUTURE

Q14 Introduction of PHQ-9 within conversation
Theme 1 Introduction Flow — Codes: INTRO-SMOOTH, INTRO-RUSHED, INTRO-PROMPTED, INTRO-MISSED
Theme 2 Conversation Disruption — Codes: DISR-NONE, DISR-CUT, DISR-PACING, DISR-UI
Theme 3 Tone/Naturalness — Codes: TONE-HUMAN, TONE-MECH

Q16 Naturalness of conversation
Theme 1 Tone — Codes: TONE-HUMAN, TONE-ROBOT, TONE-MIX
Theme 2 Conversation Flow — Codes: FLOW-SMOOTH, FLOW-DISJOINT, FLOW-REPEAT, FLOW-PACING
Theme 3 Understanding & Appropriateness — Codes: UNDERSTAND-GOOD, UNDERSTAND-MISS
Theme 4 Voice & Interface — Codes: VOICE-NAT, VOICE-ROBOT, UI-ISSUE

Q17 Handling sensitive depression topics
Theme 1 Positive Factors — Codes: A-EMPATH, A-VALIDATE, A-ADVICE, A-SAFE, A-FLOW
Theme 2 Limitations — Codes: B-ROBOT, B-SHALLOW, B-REPEAT, B-SLOW, B-RUSH, B-SELF

Q18 Comfort expressing feelings
Theme 1 Comfort Factors — Codes: CF-ANON, CF-NEUT, CF-EMPA, CF-AFFIRM, CF-SAFE, CF-GUIDE
Theme 2 Discomfort Factors — Codes: DF-PRIV, DF-ROBOT, DF-SHAL, DF-MISINT, DF-RUSH, DF-SELF, DF-NOHELP

Q19 Helpfulness of recommendations
Theme 1 Helpful Factors — Codes: HP-ACTION, HP-RESOURCE, HP-PERSONAL, HP-VALIDATE, HP-SUMMARY, HP-ACCESS, HP-SOOTHE, HP-ACCURATE
Theme 2 Unhelpful Factors — Codes: HN-GENERIC, HN-NONE, HN-REPEAT, HN-SHALLOW, HN-UNFIT, HN-LENGTH, HN-SLOW, HN-SELF

Q20 Voice clarity and tone
Theme 1 Positive Contributors — Codes: VC-CLEAR, VC-CALM, VC-MATCH, VC-HUMAN, VC-ENGAGE, VC-ACCESS
Theme 2 Negative Contributors — Codes: VN-ROBOT, VN-SLOW, VN-LONG, VN-SEQ, VN-AUTO, VN-MISMATCH, VN-CLARITY, VN-LANG, VN-TEXTPREF, VN-TIME

Q22 HopeBot compared with professional practitioners
Theme 1 Professional-like Areas — Codes: PRO-KNOW, PRO-EMPATH, PRO-STRUCT, PRO-REFLECT, PRO-ACCESS, PRO-RESOURCE, PRO-CALM
Theme 2 Limitations — Codes: LIM-HUMAN, LIM-NONVERB, LIM-PERSONAL, LIM-DEPTH, LIM-ADAPT, LIM-EMOTION, LIM-TECH, LIM-TRUST, LIM-SUB

Q23 Strengths, weaknesses, and improvements
Theme 1 Strengths — Codes: PRO-Access, PRO-Privacy, PRO-NonJudg, PRO-Struct, PRO-Know, PRO-Empath, PRO-Speed, PRO-Voice, PRO-Screen, PRO-Guide
Theme 2 Limitations — Codes: LIM-NonVerb, LIM-Depth, LIM-Personal, LIM-Empath, LIM-Robot, LIM-Slow, LIM-Tech, LIM-Security, LIM-Scope, LIM-Trust, LIM-Voice
Theme 3 Improvements — Codes: IMP-Speed, IMP-UX, IMP-Warmth, IMP-Personal, IMP-Depth, IMP-NonVerb, IMP-Privacy, IMP-Resources, IMP-VoiceOpt, IMP-Engage

Q24 Future use and recommendation
Theme 1 Willingness — Codes: PRO-USE, PRO-REC, PRO-DUAL, COND-TRY, LIM-NO, LIM-PREFALT
Theme 2 Reasons — Codes: PRO-Free, PRO-Access, PRO-Quick, PRO-Easy, LIM-Slow, PRO-Privacy, LIM-Privacy, PRO-Help, PRO-Comfort, LIM-Depth, LIM-Scope, PRO-Warm, LIM-Robot, LIM-UX, IMP-Feature, IMP-Emotion, IMP-Accuracy

Q25 Future of AI in mental health
Theme 1 Benefits — Codes: ADV-Access, ADV-Speed, ADV-EarlyDetect, ADV-Scale, ADV-Anonym, ADV-Cost, ADV-Companion, ADV-Personal, ADV-Tool
Theme 2 Risks — Codes: LIM-Accuracy, LIM-Depth, LIM-Privacy, LIM-Ethics, LIM-Replace, LIM-Sensitive, LIM-UX, LIM-Regulate, LIM-Context, LIM-Alt

Theme 3 Improvements — Codes: IMP-Improve, IMP-Interface, IMP-PrivacySafe, IMP-Supervise, IMP-Personalise, IMP-Crisis
Theme 4 Meta-stances — Codes: BAL-Mixed, UNC-Unknown

**S7 Table. Distribution of thematic codes for participant responses to open-ended survey questions (Q8–Q18).**

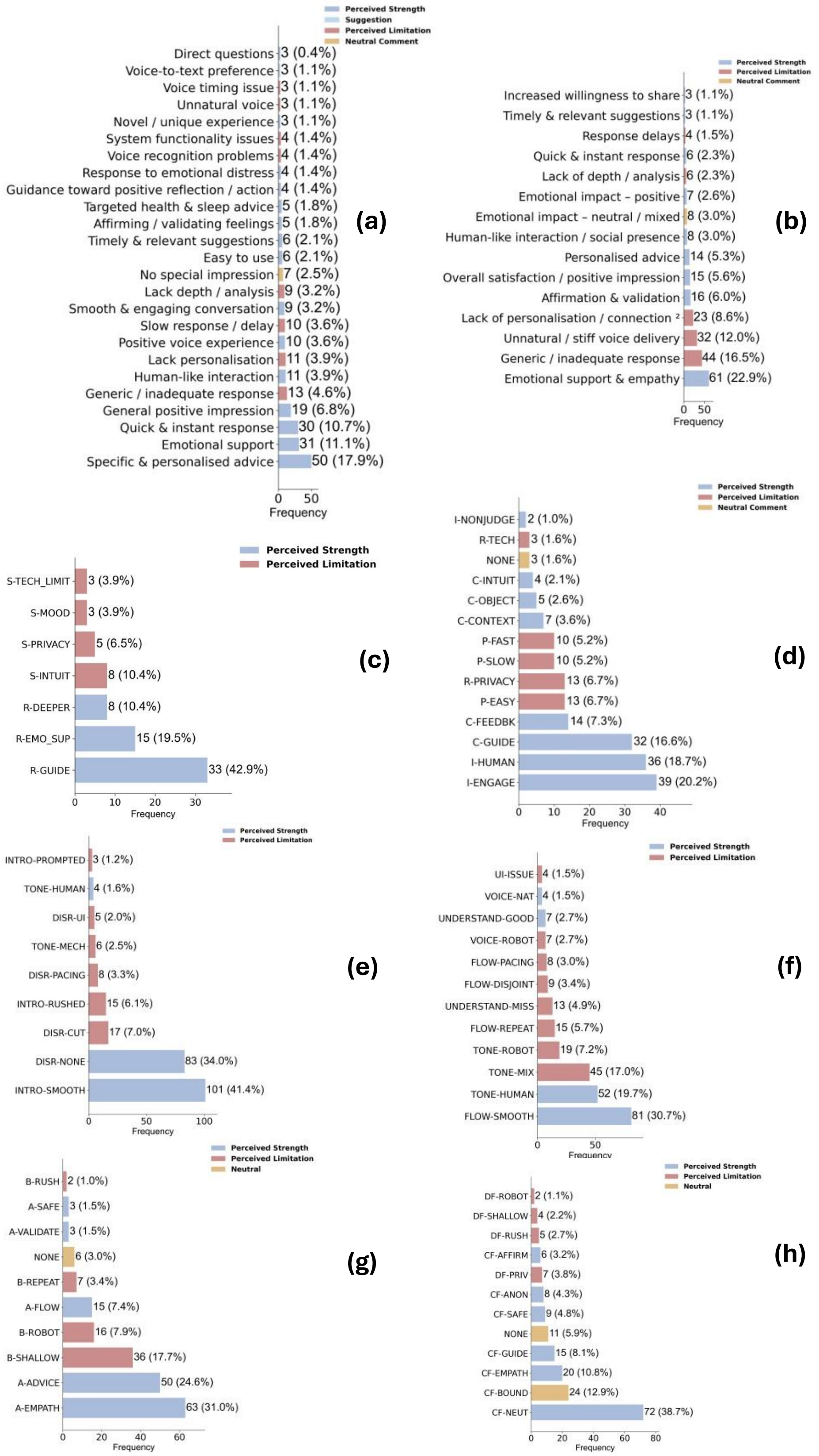

**S7 Text. Summary of coded user responses to HopeBot interaction questions (Q8–Q18).**

This composite figure visualises eight subplots (a–h), each corresponding to a specific open-ended question from the HopeBot user study. Horizontal bar charts illustrate the frequency and proportion of thematically coded responses. To enhance visual clarity, only codes that occurred at least twice ($n \geq 2$) are shown; responses coded fewer than twice were excluded from the plots but were included in percentage calculations to ensure consistency with reported results.

Bars are colour-coded to reflect the nature of user feedback:

- Perceived Strength
- Perceived Limitation
- Suggestion
- Neutral/Conditional Comment

a, Q8 – Was there anything that stood out to you during your interaction with HopeBot?
b, Q9 – Did you feel understood by HopeBot, or did it ever feel robotic?
c, Q12 – Is there a difference between your self-assessed PHQ-9 score and the HopeBot-assisted one? Which results do you find more trustworthy?
d, Q13 – How does using HopeBot to complete the PHQ-9 compare to completing it alone at home? Which did you prefer?
e, Q14 – Did HopeBot introduce the PHQ-9 naturally or disruptively?
f, Q16 – Did the overall conversation with HopeBot feel natural or robotic?
g, Q17 – How well did HopeBot handle sensitive depression-related topics (1–10)?
h, Q18 – How comfortable did you feel expressing your feelings without judgment (1–10)?

Proportions (in parentheses) were calculated based on the total number of valid responses per item (see Methods). This figure supplements the qualitative findings by illustrating the distribution and relative weight of user experiences across key domains of chatbot interaction.

**S8 Table. Distribution of thematic codes for participant responses to open-ended survey questions (Q19–Q25).**

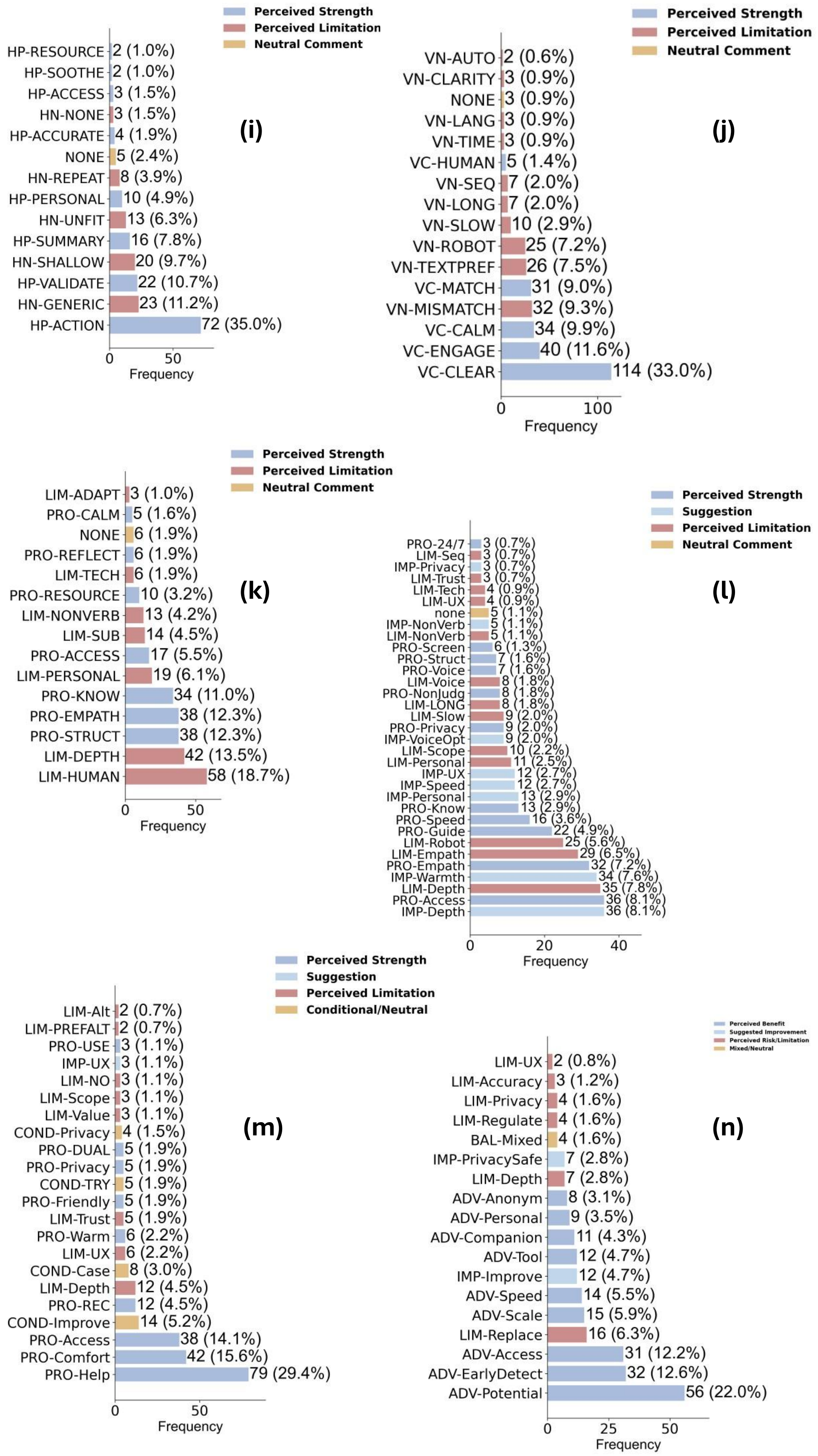

**S8 Text. Summary of coded user responses to HopeBot interaction questions (Q19–Q25).**

This composite figure presents six horizontal bar charts (i–n), each corresponding to a distinct open-ended question from the HopeBot user study (Q19–Q25). For each item, thematically coded responses are visualised in terms of frequency and percentage, with colour-coding to indicate the nature of user feedback:

- Perceived Strength
- Perceived Limitation
- Suggestion
- Neutral/Conditional Comment

To optimise clarity, only codes that occurred at least twice ($n \geq 2$) are displayed in the plots; codes with fewer than two occurrences are omitted from the bars but included in percentage calculations to ensure consistent denominators with the main results. Proportions in parentheses reflect the share of total valid responses for each question (see Methods for details). The subplots correspond to the following open-ended questions:
i, Q19 – On a scale of 1–10, how helpful were HopeBot's recommendations? What influenced your rating?
j, Q20 – On a scale of 1–10, how clear was HopeBot's voice output? Did its tone match your expectations? Were you willing to listen to the entire response? Please provide the reasoning behind your rating.
k, Q22 – In your experience using HopeBot, in what aspects do you believe its mental health support services approach the level of professional mental health practitioners? Conversely, in which areas do you find it still difficult to serve as a substitute? Please elaborate on your perspective.
l, Q23 – What were HopeBot's strengths and weaknesses, and how could it better support mental health and depression screening?
m, Q24 – Would you feel comfortable using HopeBot in the future or suggesting it to someone else? What factors influence your decision?
n, Q25 – How do you see AI and technology shaping the future of mental health screening and support?

This figure supplements the qualitative findings by illustrating the relative distribution and prominence of user-reported experiences, preferences, and suggestions across several key domains of chatbot functionality, perceived professionalism, strengths and weaknesses, future willingness, and broader perspectives on AI in mental health.